\documentclass[conference]{IEEEtran}
\ifCLASSINFOpdf
\else
\fi
\usepackage{url} 
\usepackage{flushend} 
\usepackage{graphicx} 
\usepackage{subcaption} 
\usepackage{epstopdf} 
\usepackage[cmex10]{amsmath} 
\usepackage{amssymb}
\usepackage{mathrsfs}
\usepackage{algorithm}
\usepackage{algorithmicx}
\usepackage{algpseudocode}
\usepackage{caption} 
\usepackage{tikz} 
\usetikzlibrary{snakes}

\DeclareMathOperator*{\argmin}{argmin}

\newcommand{\bheading}[1]{{\vspace{4pt}\noindent{\textbf{#1}}}}

\newcommand{\ignore}[1]{}

\algnewcommand\algorithmicinput{\textbf{INPUT:}}
\algnewcommand\INPUT{\item[\algorithmicinput]}
\algnewcommand\algorithmicoutput{\textbf{OUTPUT:}}
\algnewcommand\OUTPUT{\item[\algorithmicoutput]}

\begin{document}
%
\title{Face Flashing: a Secure Liveness Detection Protocol based on
              Light Reflections}

\author{\IEEEauthorblockN{Di Tang}
\IEEEauthorblockA{Chinese University of Hong Kong \\
td016@ie.cuhk.edu.hk}
\and
\IEEEauthorblockN{Zhe Zhou}
\IEEEauthorblockA{Fudan University\\
zhouzhe@fudan.edu.cn}
\and
\IEEEauthorblockN{Yinqian Zhang}
\IEEEauthorblockA{Ohio State University\\
yinqian@cse.ohio-state.edu}
\and
\IEEEauthorblockN{Kehuan Zhang}
\IEEEauthorblockA{Chinese University of Hong Kong\\
khzhang@ie.cuhk.edu.hk}}


%


\IEEEoverridecommandlockouts
\makeatletter\def\@IEEEpubidpullup{9\baselineskip}\makeatother
\IEEEpubid{\parbox{\columnwidth}{
    Network and Distributed Systems Security (NDSS) Symposium 2018\\
    18-21 February 2018, San Diego, CA, USA\\
    ISBN 1-1891562-49-5\\
    http://dx.doi.org/10.14722/ndss.2018.23176\\
    www.ndss-symposium.org
}
\hspace{\columnsep}\makebox[\columnwidth]{}}

\maketitle

\begin{abstract}

  Face authentication systems are becoming increasingly prevalent, especially with the rapid development of Deep Learning technologies. However, human facial information is easy to be captured and reproduced, which makes face authentication systems vulnerable to various attacks. \textit{Liveness detection} is an important defense technique to prevent such attacks, but existing solutions did not provide clear and strong security guarantees, especially in terms of time.

To overcome these limitations, we propose a new liveness detection protocol called \textit{Face Flashing} that significantly increases the bar for launching successful attacks on face authentication systems. By randomly flashing well-designed pictures on a screen and analyzing the reflected light, our protocol has leveraged physical characteristics of human faces: reflection processing at the speed of light, unique textual features, and uneven 3D shapes. Cooperating with working mechanism of the screen and digital cameras, our protocol is able to detect subtle traces left by an attacking process.

  To demonstrate the effectiveness of Face Flashing, we implemented a prototype and performed thorough evaluations with large data set collected from real-world scenarios. The results show that our Timing Verification can effectively detect the time gap between legitimate authentications and malicious cases. Our Face Verification can also differentiate 2D plane from 3D objects accurately. The overall accuracy of our liveness detection system is 98.8\%, and its robustness was evaluated in different scenarios. In the worst case, our system's accuracy decreased to a still-high 97.3\%.

\end{abstract}

\section{Introduction}
\label{sec:introduction}

User authentication is a fundamental security mechanism. However, passwords, the most widely used certificate for authentication, have widely known drawbacks in security and usability: strong passwords are difficult to memorize, whereas convenient ones provide only weak protection. Therefore, researchers have long sought alternative security certificates and methods, among which biometric authentication is a promising candidate. Biometric authentication verifies inherent factors instead of knowledge factors (e.g., passwords) and possession factors (e.g., secure tokens). Some biometric-based schemes have already been proposed. They exploit users' fingerprints, voice spectra, and irises. Face-based schemes have become increasingly widespread because of rapid developments in face recognition technologies and deep learning algorithms.


However, in contrast to other biometrics (i.e., iris and retina recognition) that are difficult for adversaries to acquire and duplicate, human faces can be easily captured and reproduced, which makes face authentication systems vulnerable to attacks. For example, adversaries could obtain numerous photographs and facial videos from social networks or stolen smartphones. Furthermore, these images can be easily utilized to build facial models of target individuals and bypass face authentication systems, benefiting from architectural advances in General-Purpose Graphics Processing Unit (GPGPU) and advanced image synthesizing techniques. Such attacks can be as simple as presenting a printed photograph, or as sophisticated as dynamically generating video streams by using video morphing techniques.


To counter such attacks, \textit{liveness detection} methods have been developed during the past decade. Crucial to such methods are challenge-response protocols, in which challenges are sent to the user who then responds in accordance with displayed instructions. The responses are subsequently captured and verified to ensure that they come from a real human being instead of being synthesized. Challenges typically adopted in studies have included blinking, reading words or numbers aloud, head movements, and handheld camera movements.

\ignore{
To defend against such attacks, \textit{liveness detection} methods have been developed during the past decade. The crucial part of these methods is a challenge-response protocol, in which challenges are sent to the user who will respond according to the instructions. The responses are then captured and
verified to ensure that they come from a real human being instead of synthesized images or videos. Typical challenges used in previous works include eye blinking, word/number reading, head movements, hand-held camera movements, etc.
}

However, these methods do not provide a strong security guarantee. Adversaries may be able to bypass them by using modern computers and technology. More specifically, as Li et al.~\cite{li2015seeing} argued, many existing methods are vulnerable to media-based facial forgery (MFF) attacks. Adversaries have even been able to bypass FaceLive, the method proposed by Li et al. and designed to defend against MFF attacks, by deliberately simulating the authentication environment.

\ignore{
Unfortunately, these methods are unable to provide a strong security guarantee. It is unclear whether adversaries could bypass them with the help of modern computers and technologies. More specifically, as mentioned by Li et
al.~\cite{li2015seeing}, many existing methods are vulnerable to
\textit{media-based facial forgery} (MFF) attacks. Even facing \textit{FaceLive}, the method proposed by Li et al. and designed to defend against MMF attacks, adversaries can bypass it, with considerable chances, by deliberately simulate the authentication environment.
}

We determined that the root cause of this vulnerability is the lack of strict time verification of a response. That is, the time required for a human to respond to a movement challenge is long and varies among individuals. Adversaries can synthesize responses faster than legitimate users by using modern hardware and advanced algorithms. Therefore, previous protocols could not detect liveness solely on the basis of response time.

\ignore{
We identify that the root cause of this drawback is the lack of strict timing verification
against response. Specifically, the time needed for a human to respond a
movement challenge is pretty long and quite variable for different people.
As a result, adversaries could synthesize responses even faster than the benign users, with the help of modern hardware and advanced algorithms. So, in previous protocols, liveness cannot be detected solely in term of time.
}

To address this vulnerability, we propose a new challenge–response protocol called Face Flashing. The core proposal of this protocol is to emit light of random colors from a liquid-crystal display (LCD) screen (our challenge) and use a camera to capture the light reflected from the face (our response). The response generation process requires negligible time, whereas forging the response would require substantially more time. By leveraging this substantial difference, Face Flashing thus provides effective security in terms of time.

\ignore{
To address this issue,
we propose a new challenge-response protocol called
\textbf{Face Flashing}. The key idea is to emit light with random colors from an LCD
screen (our challenges), and use a camera to capture the light
reflected from the face (our responses). The response generation process needs no time while forging the response needs prominently more time. Therefore, leveraging this substantial difference, Face Flashing can provide strong security guarantee in terms of time.
}

The security of the Face Flashing protocol is based on two factors: time and shape. We use linear regression models and a neural network model to verify each factor, respectively. Our verification of time ensures that the response has not been falsified, whereas verification of shape ensures that the face shape is stereo and face-like. By using these two verifications, our protocol simultaneously satisfies the three essentials of a secure liveness detection protocol. First, we leverage an unpredictable challenge, flashing a sequence of effectively designed, randomly generated images. Second, our responses are difficult to forge not only because of the difference in time but also in the effort required to generate responses. In particular, legitimate users need not perform any extra steps, and legitimate responses are generated automatically (through light reflection) and instantaneously, whereas adversaries must expend substantially more effort to synthesize quality responses to bypass our system. Third, we can effectively verify the genuineness of responses by using our challenges. Specifically, we verify users on the basis of the received responses; for example, by checking whether the shiny area in the response accords with the challenge (lighting area in challenges will always produce highly intensive responses in a local area). The detailed security analysis and our adversary model are presented in later sections of this paper.

\ignore{
The security guarantee of Face Flashing is built on two terms: \textbf{time} and \textbf{shape}. We use linear regression models and a neural network model to do verifications on them respectively. Our timing verification ensures the response has not been falsified and face verification ensures the shape is stereo and face-like. With these two verifications, 
our protocol simultaneously satisfies three essentials required by a secure liveness detection protocol. 
First, we leverage unpredictable challenges, flashing a sequence of well-designed images that are randomly generated.
Second, our responses are hard to be forged, because the difference not only in time but also in the amount of efforts needed to generate responses. Particularly, for benign users, they do not need to do anything extra and legitimate responses are generated automatically (light reflection) without any delay, while for adversaries, they need significantly more efforts to synthesize quality responses that can pass our verifications.
Third, we can effectively check the genuineness of responses according to our challenges. Specifically, we verify it on the challenges recovered from received responses, e.g., checking whether the shiny area in the response accords with the challenge (lighting area in challenges will always produce highly intensive responses in a local area). The detailed security analysis and our adversary model will be presented in later parts.
}

\textbf{Contributions.}
Our paper's contributions are three-fold:
\begin{itemize}

\item \textbf{A new liveness detection protocol, Face Flashing.}
We propose Face Flashing, a new liveness detection protocol, that flashes randomly generated colors and verifies the reflected light. In our system, adversaries do not have the time required to forge responses during authentication.

\item \textbf{Effective and efficient verifications of timing and face.} 
By employing working mechanisms of screens and digital cameras, we design a method that uses linear regression models to verify the time. Furthermore, by using a well-designed neural network model, our method verifies the face shape. By combining these two verification procedures, our protocol provides strong security.

\item \textbf{Implementation of a prototype and evaluations.} 
We implement a prototype and conduct thorough evaluations. The evaluation results suggest that our method performs reliably in different settings and is highly accurate.

\end{itemize}

\textbf{Roadmap.} This paper is organized as follows:
Section~\ref{sec:background}
introduces the background and Section~\ref{sec:adversaryModel} describes our adversary model and preset assumptions. Section~\ref{sec:design} details the design of our protocol.
Section~\ref{sec:securityAnalysis} presents the security analysis.
Section~\ref{sec:evaluation} elucidates experiment settings and evaluation
results.  Section~\ref{sec:related} summarizes related works.
Section~\ref{sec:discussion} and Section~\ref{sec:limitation} discusses limitations and future works.
Finally, Section~\ref{sec:conclusion} concludes this paper.

\section{Background}
\label{sec:background}

In this section, we describe the typical architecture of face-based authentication
systems (Section \ref{sec:systemArchitecture}). Subsequently, we briefly review attacks and solutions of liveness detection (Section \ref{sec:attackingTree}). 

\subsection{Architecture of Face Authentication Systems}
\label{sec:systemArchitecture}

\begin{figure}[tbh]
\centering
\includegraphics[width=0.49\textwidth]{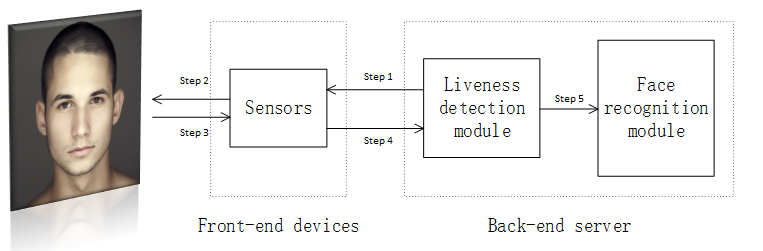}
\caption{A typical face authentication system.}
\label{fig:architecture}
\end{figure}

A typical architecture of face authentication system is illustrated in
Fig~\ref{fig:architecture}. It is divided into two parts: front-end devices and
the back-end server. The front-end devices comprises camera and auxiliary sensors such as flash lamps, microphones. The back-end server contains two main modules: a liveness detection module and a face recognition module. When
the user commences the authentication process, the liveness detection module is initiated and sends generated parameters to front-end devices (Step 1). Subsequently, the front-end devices synthesize challenges according to the received parameters and deliver them to the user (Step 2). After receiving the challenges, the user makes expressions, such as smiling blinking, as responses. The sensors in the front-end devices capture such responses and encode them (Step 3). Either in real time or in post processing, the front-end devices send the captured responses to the liveness detection module in the back-end server (Step 4). The liveness detection module gathers all decoded data and checks whether the user is an actual human being. If so, the liveness detection module selects some faces among all the responses and delivers them to the face recognition module to determine the identity of the user (Step 5).

\subsection{Attacks and Solutions on Liveness Detection}
\label{sec:attackingTree}

\begin{figure}[tbh]
\centering
\includegraphics[width=0.49\textwidth]{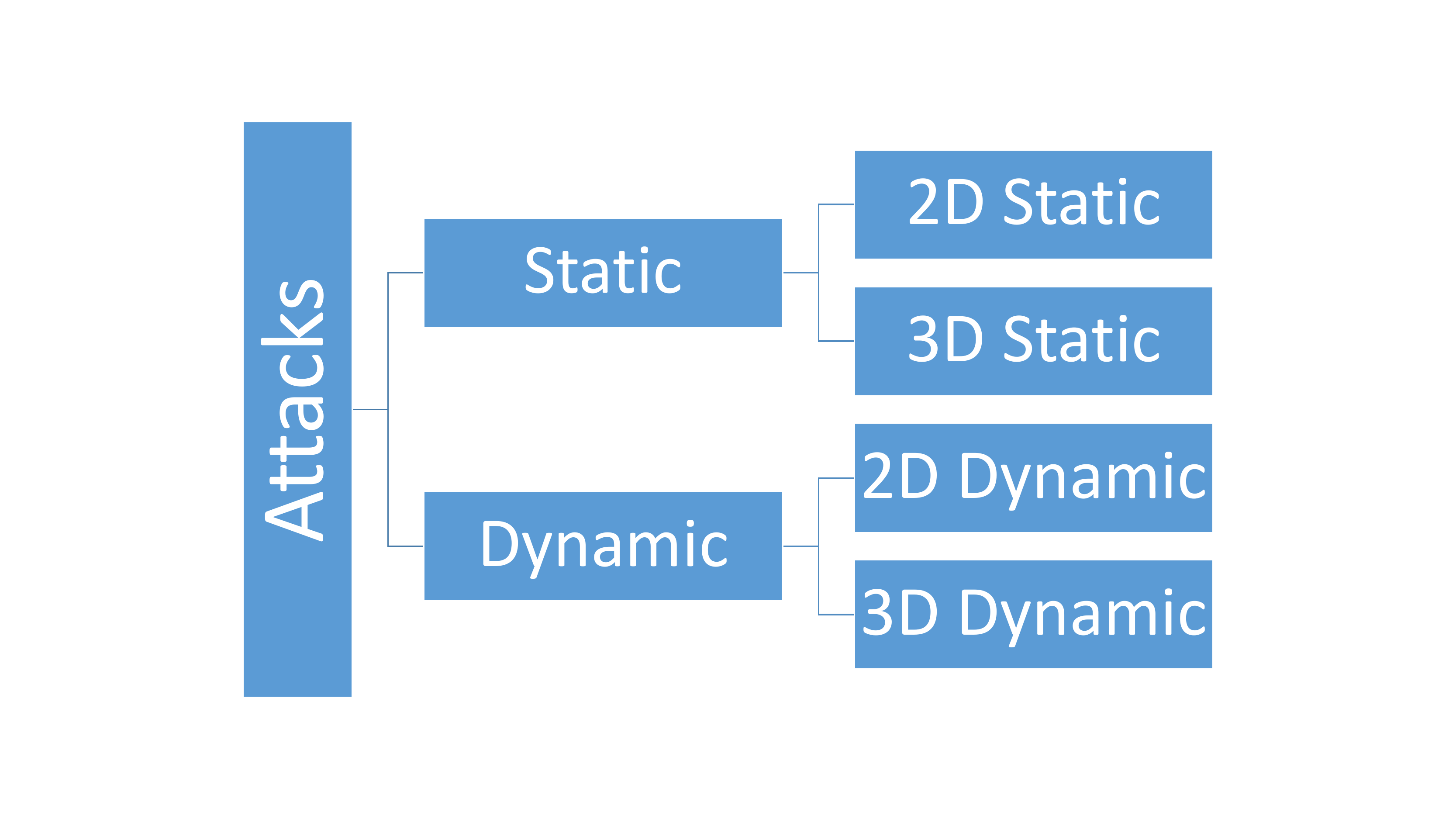}
\caption{The attacking tree.}
\label{fig:attackingtree}
\vspace{-0.1in}
\end{figure}

In recent years, plenty of attacks have been developed to exploit the flaw that face recognition algorithms cannot determine whether a photograph taken by the front-end camera is captured from a real face, even if the recognition accuracy of some has exceeded human beings. 
In this study, we divide attacks into four categories and organize them as a tree, known as the attack tree, which is displayed in Fig~\ref{fig:attackingtree}. We first separate attacks into two categories: \textit{static} and \textit{dynamic}. \textit{Static} attacks refer to the use of static objects, such as photographs, plastic masks, and paper, as well as transformations of these objects (e.g., folding, creating holes through them, and assembling them into certain shapes). Attacks using dynamic or partially dynamic objects are categorized into dynamic branch. Subsequently, we separate attacks into four subcategories: two-dimensional (2D) static, three-dimensional (3D) static, 2D dynamic, and 3D dynamic. The 3D branches refer to attacks that use stereo objects, including sculptures, silicone masks, and robots. More precisely, these objects must have notable stereo characters of human faces, such as a prominent nose, concave eye sockets and salient cheekbones; otherwise, the attacks are categorized into \textit{2D} branches. Organized by this attacking tree, a brief review of relative attacks and solutions is presented below.

In the \textit{2D static} branch, photograph-based attacks are the predominant form of attack.
They are easily launched and effective at compromising primary recognition algorithms. Tracking eye movement was the first method proposed to counter such attacks. Jee et al.~\cite{jee2006liveness} proposed a method for detecting eyes in a sequence of facial images. Variations around the eyes are calculated, and whether the face is real is determined. Their basic assumption is that blinking and the uncontrolled movements of pupils are human nature behaviors. However, this method cloud be compromised by an adversary wearing a mask with eyeholes. A similar idea exploiting Conditional Random Fields (CRFs) was proposed by Sun et al.~\cite{sun2007blinking}. Its limitation is the same. Subsequently, lip movement and lip reading methods were developed by Kollreider et al.~\cite{kollreider2007real} for liveness detection. However, their method can also be fooled using carefully transformed photographs.

\ignore{
To relief this threat, checking the movement of eyes is first proposed. Hyung-Keun Jee et al.~\cite{jee2006liveness} proposed a method to detect eyes in a sequence of facial images. Varying around eyes is calculated and whether the face is real or not is determined. Their basic assumption is that blinking and the uncontrolled movements of pupils are human natures. Even though the idea is good, their method can be still compromised by an adversary who masked by a simple photo with 2 holes. Another similar idea exploiting Conditional Random Fields (CRFs) was published by Lin Sun et al.~\cite{sun2007blinking}. Naturally, its limitation is the same. Besides, lip movement and lip reading were developed by Kollreider et al.~\cite{kollreider2007real} for liveness detection. However, similarly, their method can also be fooled by carefully designed transformations of photos. 
}

To distinguish faces from photographs, Li et al.~\cite{li2004live} leveraged the fact that faces and photographs have different appearances in frequency space. They conducted 2D Fourier spectral analysis to extract the high-frequency components of input images; faces contain more components in the high-frequency region than do photographs. However, adversaries can print a high-resolution photograph to bypass this method. Jukka et al.~\cite{maatta2011face} observed that printed photographs are pixelized. That is, a face has more details than a photograph. Thus, they used a support vector machine (SVM) to extract microtextures from the input image. Later, Kim et al.~\cite{kim2012face} leveraged a more powerful texture descriptor, Local Binary Pattern (LBP) to enhance performance. They additionally analyzed the information residing in both the low- and high-frequency regions. However, all these types of solutions have a common drawback: low robustness. Motion blur or noise from the environment impairs their performance. Moreover, these methods are useless against screen-based attacks~\cite{chakraborty2014overview}.

\ignore{
To distinguishing real faces from photos, Li et al.~\cite{li2004live} leveraged the fact: photos and faces have different views in frequency space. They performed 2D Fourier Spectra analysis to extract high-frequency components of input images, as the real face produces more components than the photos in the high-frequency region. 
However, adversaries can print high-resolution photos to bypass this method. On the other hand, 
Jukka et al.~\cite{maatta2011face} observed that printed photos are pixelized. That is to say a real face has more details than the fake one. Armed with it, they used Support Vector Machine (SVM) to extract micro-texture from the input image.
Later on, Gahyun Kim et al.~\cite{kim2012face} leveraged a more powerful texture descriptor, Local Binary Pattern (LBP), to enhance the performance. Further, they analyzed the information residing in both the low-frequency regions and the high-frequency regions. 
But, all these solutions have a common drawback, low robustness. Motion blur or environment noise will fade their performance. Worse, these methods are useless to defeat screen-based attacks~\cite{chakraborty2014overview}.
}

Because of strategies designed to protect against 2D attacks, adversaries have attempted to exploit 3D objects in 3D static attacks. Thus, researchers have developed novel methods to defend against these attacks also. Lagorio, et al.,~\cite{lagorio2013liveness} proposed a method to detect 3D facial features. They employed a two- 
camera device to ascertain the surface curvature of the subject seeking authentication. If the surface curvature is low, the subject is judged to be malicious. Although the accuracy is almost 100\%, the sophisticated device required is expensive and the computational cost is unacceptable. Furthermore, Wang, et al.,~\cite{wang2013face} leveraged a one-camera, face alignment algorithm to ascertain 3D shape on the basis that forged faces are usually flatter than real faces. However, this method performed unsatisfactorily when applied to spectacle-wearers because of the limited capability of the face alignment algorithm.

\ignore{
Along with emerging of mitigations against 2D attacks, adversaries tried to exploit 3D objects to launch \textit{3D static} attacks. Meanwhile, corresponding defenses have also arisen. 
An approach detecting 3D features of faces was proposed by Andrea Lagorio et al.~\cite{lagorio2013liveness}. They leveraged a 2-camera device to recover the surface curvature of the subject under authentication. If an object has low curvature, it will be judged as a malicious one. Even though the accuracy is as high as almost 100 percent, but neither the price of the sophisticated device nor the computational cost is acceptable. Further, Wang et al.~\cite{wang2013face} leveraged 1-camera and face alignment algorithm to recover the 3D shape. Their logic is that forged faces are usually flatter than real faces. But limited by the capability of face alignment algorithm, the performance of this method is low when facing someone who wears glasses.
}

In response to technological developments, adversaries must in turn develop more sophisticated attacks. One method is pasting stereo materials onto photographs. In contrast, researchers have developed practical and efficient methods to counter these threats, with the help from developments in computer vision. The fundamental idea behind these methods is that adversaries cannot manipulate static objects to simulate instructed expressions, even if these objects are similar to the human face. Thus, a common challenge-response protocol has been adopted, whereby users are asked to make expressions as the instructions, including happiness, despair, and surprise. Such systems subsequently compare the captured video with stored data.

\ignore{
As the bar is raising, adversaries have to develop more sophisticated attacks. A striking attack is pasting various stereo materials onto photos. On the opposite, defenders developed some practical and efficient methods to relieve such threats, simulating by exploration of developments in Computer Vision. The fundamental idea embedded in these methods is that adversaries can hardly steer a static object to make instructed expressions, even if this object is similar with the human face. Armed with this idea, a prevalent challenge-response protocol has been proposed. It instructs the user make some expressions like smiling, crying, surprising, etc., and compares the captured video with previously stored data. 
}

However, more powerful \textit{2D dynamic} attacks have been developed, in which adversaries have exploited advanced deep learning models and personal computers with powerful processors. These attacks work by merging a victim's facial characteristics with photographs of the victim and using these forged photographs to bypass the face recognition algorithm. Furthermore, even if this operation requires time, adversaries can prepare photographs beforehand and launch offline attacks, sending forged photographs to an online authentication system.

\ignore{
However, more powerful \textit{2D dynamic} attack have been developed, in which adversaries have exploited advanced deep learning models and powerful personal computer. Particularly, they merge victim's facial characters onto photos of themselves and use these forged photos to fool the face recognition algorithm. Worse, even if this operation needs time, adversaries can complete it beforehand and launch offline attacks, sending their previously forged responses to an online authentication system. 
}

To counter these new 2D dynamic threats, some solutions have been proposed. Bao et al..~\cite{bao2009liveness} introduced a method based on optical flow technology. The authors found that the motions of 2D planes and 3D objects in an optical flow field are the same in translation, rotation, and moving, but not in swing. They used this difference to identify fake faces. Unfortunately, two drawbacks undermine this method: first, an uneven object will fail it; second, the method does not consider variation in illumination. Kollreider, et al.,~\cite{kollreider2009non} developed a method for detecting the positions and velocities of facial components by using model-based Gabor feature classification and optical flow pattern matching. However, its performance was impaired when keen edges were present on the face (e.g., spectacles or a beard). The authors admitted that the system is only error-free if the data contain only horizontal movements. Findling et al.~\cite{findling2012towards} achieved liveness detection by combining camera images and movement sensor data. They captured multiple views of a face while a smartphone was moved. Li et al.~\cite{li2015seeing} measured the consistency in device movement detected using inertial sensors and changes in the perspective of a head captured on video. However, both methods were demonstrated to be compromised by Xu et al.~\cite{xu2016virtual}, who constructed virtual models of victims on the basis of publicly available photographs of them.

\ignore{
To relief these newly \textit{2D dynamic} threats, some solutions were published. Bao et al.~\cite{bao2009liveness} introduced a method based on optical flow technology. The authors found motions of 2D planes and 3D objects in optical flow field are the same in translation, rotation, moving scenes, but swing. They detected this difference to distinguish the fake face. Unfortunately, there are 2 main drawbacks of this method: first, uneven objects will fail it; second, it is not robust to variation of illumination. 
Kollreider et al.~\cite{kollreider2009non} developed a method to detect different positions and velocities of facial parts with the help of model-based Gabor feature classification and optical flow pattern matching. However, their performance is faded by keen edges on the face like glasses and beard. And it is mentioned by authors that the system will be error-free if the data only contains horizontal movements. Findling et al.~\cite{findling2012towards} achieved liveness detection by combining camera images and movement sensor data. They captured multiple face views while moving the smartphone. A more comprehensive work is done by Li et al.~\cite{li2015seeing}. They measured the consistency of device movement detected from inertial sensors and the changes of head pose captured from facial video. But both 2 methods were compromised by Xu et al.~\cite{xu2016virtual} who built the virtual model of victims from their public photos.
}

Less adversaries have attempted to launch \textit{3D dynamic} attacks. They can reconstruct a 3D model of a victim's face~\cite{xu2016virtual} in virtual settings but hardly fabricate them in real scenes. We illustrate the difficulties in launching 3D dynamic attacks using the following three examples: First, building a flexible screen that can be molded into the shape of a face is expensive and may fail because the reflectance from a screen differs from that of a face. Second, 3D printing a ”soft” mask is impractical, being limited by the printing materials available (see Section~\ref{sec:secureAttack} for a fuller explanation). Third, building an android is infeasible and intricate and would involve face animation, precision control, and skin fabrication. Additionally, building an android is costly, particularly a delicate android “face.”

\ignore{
As for launching \textit{3D dynamic} attacks, adversaries have not marched into this field, according to our knowledge. Actually, they can reconstruct the 3D model of victim's face~\cite{xu2016virtual} in virtual scenes, but can hardly fabricate it out in real scenes. We demonstrate their difficulties in three possible methods. First, building a so flexible screen that can be shaped to be a face is expensive and may fail, because the reflectance of the screen is different from the face. Second, 3D printing a "soft" mask is impractical, limited by the materials used to print (will be explained in Section~\ref{sec:secureAttack}). Third, building a humanoid is herculean and intricate, which involves face animation, precision control, skin fabrication, etc. Besides, building a robot is costly, especially a delicate "face".   
}

On the basis of the above discussion, we observe that the current threats are principally 2D dynamic attacks because static attacks have been effectively neutralized and 3D dynamic attacks are hard to launch.

\ignore{
Based on above knowledge, we observe that the current threats mainly come from \textit{2D dynamic} branch, as attacks from the \textit{static} branch are blocked and it is hard to carry out \textit{3D dynamic} attacks.
}

\section{Adversary Model}
\label{sec:adversaryModel}

In this section, we present our proposed adversary model and assumptions.

We assume adversaries' goal is to bypass the face authentication systems by impersonating victims, and the objective of our proposed methods is to raise the bar for such successful attacks significantly. As will be demonstrated in the limitation part (section~\ref{sec:limitation}), powerful adversaries could bypass our security system, but the cost would be much higher than is currently the case. Particularly, they need to purchase or build special devices that can do all of the following operations within the period when the camera scanning a single row: (1) capture and recognize the randomized challenges, (2) forge responses depending on the random challenges, and (3) present the forged responses. For this, adversaries require high-speed cameras, powerful computers, high-speed I/Os, and a specialized screen with fast refreshing rate, etc. Therefore, it is difficult to attack our system.

Adversaries can also launch \textit{3D dynamic} attacks, such as undergoing cosmetic surgery, disguising their faces, or coercing a victim’s twin into assisting them. However, launching a successful \textit{3D dynamic} attack is much more difficult than using existing methods of MFF attack; crucially, identifying such an attack would be challenging even for humans and would constitute a Turing Test problem, which is beyond the scope of this paper. But in either case, our original goal is achieved by having increased the bar for successful attacks significantly.

\ignore{
Adversaries can also use \textit{3D dynamic} attacking techniques, like changing or disguising one's face, attacking from twins, etc. However, launch a successful \textit{3D dynamic} attack has a much higher bar than existing \textit{MMF} attacks, and more importantly, defeating these attacks itself is also a challenging problem even for human beings, which actually can be reduced to a \textit{Turing Test} problem and is definitely out of the scope of this paper. But in either case, our original goal is achieved by having increased the bar for successful attacks significantly.
}

Our method relies on the integrity of front-end devices; that is, that the camera and the hosting system that presents random challenges and captures responses have not been compromised. If this cannot be guaranteed, adversaries could learn the algorithm used to generate random challenges and generate fake but correct responses beforehand, thus undermining our system. We believe that assuming the integrity of front-end devices is reasonable in real-world settings, considering that in many places the front-end devices can be effectively protected and their integrity guaranteed (e.g., ATMs and door access controls). We cannot assume or rely on the integrity of smartphones, however. Our proposed techniques are general and can easily be deployed on different hardware platforms, including but not limited to smartphones. For simplicity, we choose to build a prototype and conduct evaluations on smartphones, but this is only for demonstration purposes. If the integrity of a smartphone can be guaranteed, by using a trusted platform module or Samsung KNOX hardware assistance, for example, our techniques can be deployed on them; otherwise this should be avoided and the proposed techniques are not tied to smartphones.

\ignore{
Besides, our method requires the integrity of front-end devices, i.e., the camera and its hosting system that presents random challenges and captures responses have not been compromised. If this cannot be guaranteed, adversaries could learn the secret used to generate random challenges and generate fake but correct responses ahead of time, and thus defeat our proposed techniques. We believe this assumption is also reasonable in real-world settings, considering that in many places the front-end devices can be well protected and its integrity can be guaranteed, such as bank ATMs, door access controls, and so on. We do NOT assume or rely on the integrity of smartphones. Our proposed techniques are general and can be implemented on or deployed to different hardware platforms easily, including but never limited to smartphones. For simplicity, we choose to build a prototype and conducted evaluations on smartphones, but this is only for demonstration purpose. If the integrity of a smartphone can be guaranteed, in some cases with hardware assistance like TPM or Samsung KNOX, our techniques then can be deployed there, otherwise it should and can be avoided, as the proposed techniques are not tied to smartphones. 
}

\section{Face Flashing}
\label{sec:design}

Face Flashing is a challenge-response protocol designed for liveness detection. In this section, we elaborate its detailed processes and key techniques to leverage flashing and reflection.

\subsection{Protocol Processes}
\label{protocolDesign}

\begin{figure}[tb]
\centering
\includegraphics[width=0.5\textwidth]{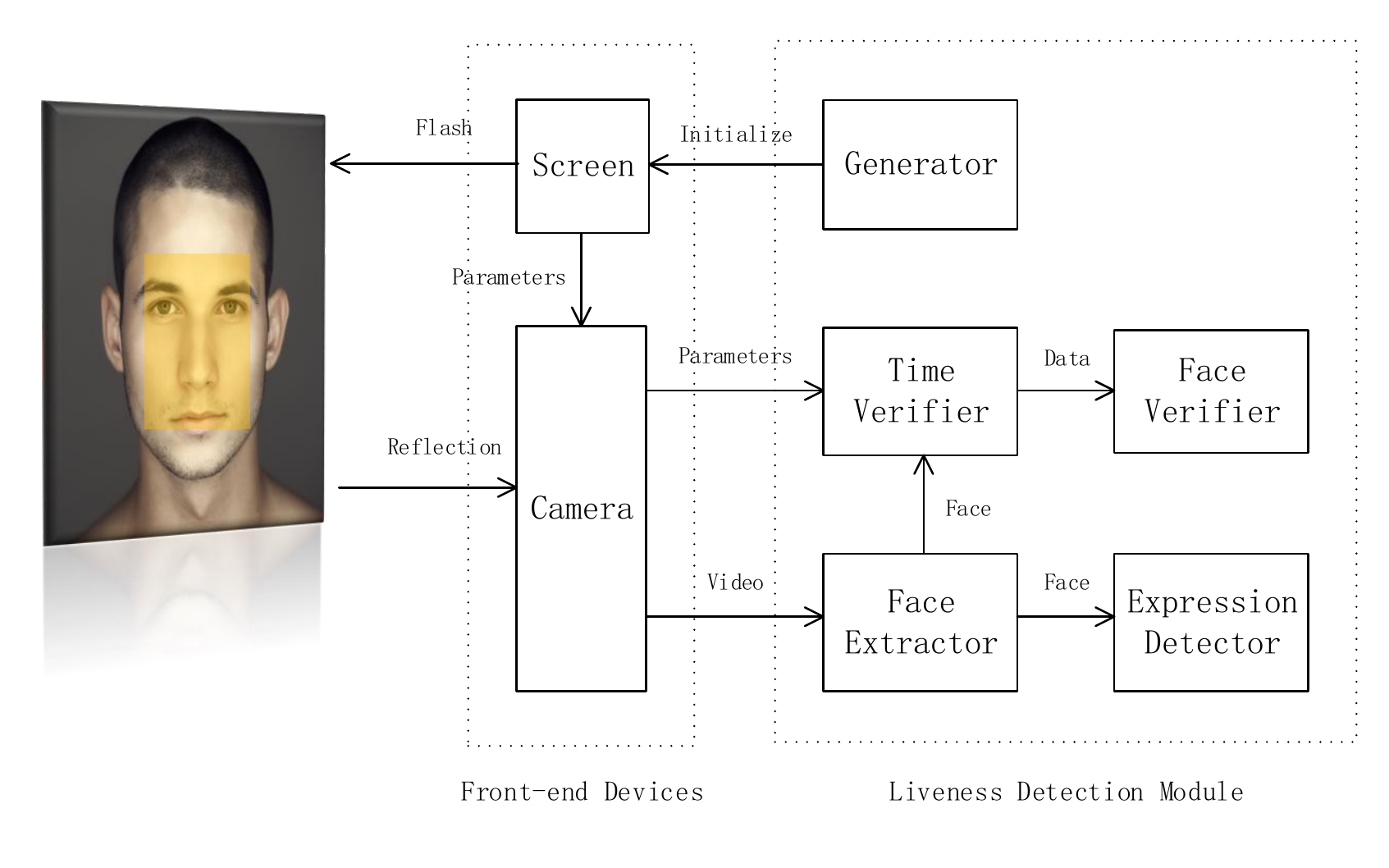}
\caption{Architecture of Face Flashing.}
\label{fig:workingFlow}
\vspace{-0.1in}
\end{figure}

The proposed protocol contains seven components, which are illustrated in Fig~\ref{fig:workingFlow}, and eight steps are required to complete once challenge-response procedure where the challenge is flashing light and the response is the light reflected from the face.

\begin{itemize}

  \item \textbf{Step 1: Generation of parameters.} 
  Parameters are produced by the \textit{Generator} of the \textit{Liveness Detection Module} running on the back-end server, which works closely with the front-end devices. Such parameters include seed and N. Seed controls how random challenges are generated, and N determines the total number of challenges to be used. Communication of parameters between the back-end server and front-end devices is protected by standard protocols such as HTTP and TLS.
  
  \ignore{
  Parameters are produced by \textit{Generator} of \textit{Liveness Detection Module} running on the back-end server that works closely with front-end devices. Such parameters includes $seed$ and $N$. $Seed$ controls how random challenges would be generated, and $N$ determines the total number of challenges to be presented. Note that communication of parameters between the back-end server and front-end devices is protected by standard protocols like HTTPS and TLS. }

  \item \textbf{Step 2: Initialization of front-end devices.} 
  After receiving the required parameters, the front-end devices initialize their internal data structures and start to capture videos from the subject being authenticated by turning on the \textit{Camera}.

  \item \textbf{Step 3: Presentation of a challenge.} 
  Once initialized, the front-end devices begin to generate challenges according to the received parameters. Essentially, a challenge is a picture displayed on a screen during one refresh period; light is emitted by the screen onto the subject’s face. The challenge can be of two types: a background challenge, which displays a pure color, and a lighting challenge, which displays a lit area over the background color. More details are specified in subsequent sections.
  
  \ignore{
  Once initialized, front-end devices begin to generate challenges according to received parameters. Essentially, a challenge is to display a picture on the \textit{Screen} during one refreshing period. The challenge can be 2 types: the background challenge showing a pure color, and the lighting challenge showing a lighting area over the background color. More details will be given later.
  }

  \item \textbf{Step 4: Collection of response.} 
  The response is the light that is reflected immediately by the subject’s face. We collect the response through the camera that has already been activated (in Step 2).
  
  \ignore{
  Response is the light that is emitted from \textit{Screen} and reflected immediately from the subject's face. We collect the response through \textit{Camera} since Step 2 where the \textit{Camera} has already been turned on.
   }
  
  \item \textbf{Step 5: Repetition of challenge-response.} Our protocol repeats Step 3 and 4 for $N$ times. This repetition is designed to collect enough responses to ensure high robustness and security so that legitimate users always pass whereas adversaries are blocked even if they accidentally guess out the correct challenges beforehand.
  
  \item \textbf{Step 6: Timing verification.} 
  Timing is the most crucial security guarantee provided by our protocol and is the fundamental distinction between genuine and fake responses. Genuine responses are the light reflected from a human face and are generated through a physical process that occurs simultaneously over all points and at the speed of light (i.e., zero delay). Counterfeit responses, however, would be calculated and presented sequentially, pixel by pixel (or row by row), through one or more pipelines. Thus, counterfeit responses would result in detectable delays. We detect delays among all the responses to verify their integrity.

  \ignore{
  Timing is the most important security guarantee provided by our protocol, and is the fundamental distinction between genuine responses and counterfeit responses. Concretely, in genuine responses, lights are reflected from a real human face, which is a physical process that is very fast; 
but counterfeit responses are calculated and presented sequentially, pixel by pixel (or row by row), through one or more pipelines. So counterfeit responses will result in detectable delays with high probability. 
}

  \item \textbf{Step 7: Face verification.} The legality of the face is verified by leveraging neural network that incorporates with both the shape and textual characters extracted from the face. This verification is necessary because without it our protocol is insufficiently strong to prevent from MFF attacks, and face verification prolongs the time required by adversaries to forge a response, which makes the difference from benign response more obvious. The details are provided in Section~\ref{sec:keyTechnique}.
  
  \item \textbf{Step 8: Expression verification.} 
  The ability to make expressions indicates liveness. We verify this ability by ascertaining whether the detected expression is the one requested. Specifically, technology from~\cite{smith2010facial} is embedded in our prototype for detecting and recognizing human expressions.
  
  \ignore{
  Making expression is an ability standing for liveness. We verify this ability by checking whether the detected expression is the one what we have already instructed. Specifically, a previous work~\cite{smith2010facial} is embedded in our prototype to detect and recognize human expressions. 
  }

\end{itemize}

Details of Step 8 are omitted in this paper so that we can focus on our two crucial steps: timing and face verifications. However, expression detection has been satisfactorily developed and is critical to our focus. Additionally, Step 8 is indispensable because it integrates our security boundary, which is elucidated in Section~\ref{sec:securityAnalysis}. The face extraction detailed in the next section is designed so that our two verification techniques are compatible with this expression detection.

\ignore{
Details of Step 8 will be omitted to avoid distraction from our two crucial steps, verifications on time and face. Actually, expression detection is well-developed and orthogonal to our focus. However, Step 8 is indispensable, as it integrates our security boundary, which will be elaborated in Section~\ref{sec:securityAnalysis}. And the Face Extraction demonstrated in next section makes our two verifications can easily cooperate with it.   
}

\subsection{Key Techniques}
\label{sec:keyTechnique}

The security guarantees of our proposed protocol are built on the timing as well as the unique features extracted from the reflected lights. In the followings, we will first introduce the model of light reflection, then our algorithm for extracting faces from video frames, and verifications on time and face.

\subsubsection{\textbf{Model of Light Reflection}}
\label{sec:reflectionModel}

Consider an image $I_{rgb} = \{I_r, I_g, I_b\}$ that is taken from a linear RGB color camera with black level corrected and saturated pixels removed. The value of $I_c, c \in \{r,g,b\}$ for a Lambertian surface at pixel position $x$ is equal to the integral of the product of the illuminant spectral power distribution $E(x,\lambda)$, the reflectance $R(x,\lambda)$ and the sensor response function $S_c(\lambda)$:
\begin{equation}
\notag
I_c(x) = \int_\Omega E(x,\lambda)R(x,\lambda)S_c(\lambda) d\lambda, c \in \{r,g,b\}
\end{equation}
where $\lambda$ is the wavelength, and $\Omega$ is the wavelength range of  all visible spectrum supported by camera sensor. From the Von Kries coefficient law~\cite{brainard1986analysis}, a simplified diagonal model is given by:
\begin{equation}
\notag
I_c = E_c \times R_c, c \in \{r,g,b\}
\end{equation}

Exploiting this model, by controlling the outside illuminant $E$, we can get the reflectance of the object. Specifically, when $E_c$ for $x$ and $y$ are the same, then 
\begin{equation}
  \frac{I_c(x)}{I_c(y)} = \frac{R_c(x)}{R_c(y)}, c \in \{r,g,b\}
\label{fa:geometry}
\end{equation} 
This means the lights captured by camera sensor at two different pixels $x$ and $y$ are proportional to the reflectance of that two pixels.

Similarly, for the same pixel point $x$, if applying two different illuminant lights $E_{c1}$ and $E_{c2}$, then:
\begin{equation}
  \frac{I_{c1}(x)}{I_{c2}(x)} = \frac{E_{c1}(x)}{E_{c2}(x)}, c1,c2 \in \{r, g, b\}
\label{fa:time}
\end{equation}
In other words, the reflected light captured by the camera in a certain pixel is proportional to the incoming light of the same pixel.

\textbf{Implications of above equations.}
Eq.(\ref{fa:geometry}) and Eq.(\ref{fa:time}) are simple but powerful. They are the foundations of our liveness detection protocols. Eq.(\ref{fa:geometry}) allows us to derive relative reflectance for two different pixels from the proportion of captured light from these two pixels. The reflectance is determined by the characteristics of the human face, including its texture and 3D shape. Leveraging Eq.(\ref{fa:geometry}), we can extract these characteristics from the captured pixels and further feed them to a neural network to determine how similar the subject's face is to a real human face.

Eq.(\ref{fa:time}) states that for a given position, when the incoming light changes, the reflected light captured by the camera changes proportionally, and crucially, such changes can be regarded as ``simultaneously'' to the emission of the incoming light because light reflection occurs at the speed of light. Leveraging Eq.(\ref{fa:time}), we can infer the challenge from the current received response and detect whether a delay occurs between the response and the challenge.

\subsubsection{\textbf{Face Extraction}}
\label{sec:faceExtraction}

To do our verifications, we need to locate the face and extract it. Furthermore, our verifications must be performed on regularized faces where pixels in different frames with the same coordinate represent the same point on the face. Concretely, when a user's face is performing expressions as instructed, it produces head movements and hand tremors. Thus, using only face detection technology is insufficient; we must also employ a face alignment algorithm that ascertains the location of every landmark on the face and neutralizes the impacts from movements. Using the alignment results, we can regularize the frames as we desired, and the regularized frames also ensure that our verifications are compatible with the expression detector.

First, We designed Algorithm 1 to quickly extract the face rectangle from every frame. In Algorithm~1, $track(\cdot)$ is our face tracking algorithm~\cite{fernando2007identification}.
It uses the current frame as the input and employs previously stored frames and face rectangles to estimate the location of the face rectangle in the current frame. The algorithm outputs the estimated rectangle and a confidence degree, $\rho$. When it is small ($\rho < 0.6$), we regard the estimated rectangle as unreliable and subsequently use $detect(\cdot)$, our face detection algorithm~\cite{yang2015convolutional}, to redetect the face and ascertain its location. We employ this iterative process because the face detection algorithm is precise but slow, whereas the face tracking algorithm is fast but may lose track of the face. Additionally, the face tracking algorithm is used to obtain the transformation relationship between faces in adjacent frames, which facilitates our evaluation of robustness (Sec~\ref{sec:val:robust}).

\begin{algorithm}
\caption{Algorithm to extract the face.}
\begin{algorithmic}[1]
\INPUT $Video$
\OUTPUT $\{F_j\}$
\For{$frame$ in $Video$}
\State $Rect, \rho$ = track($frame$)
\If{$Rect = \emptyset$ or $\rho < 0.6$}
\State $Rect$ = detect($frame$)
\State $Rect$ $\to$ track($\cdot$)
\EndIf
\State $F_j$ = $frame(Rect)$
\EndFor
\end{algorithmic}
\label{alg:extract}
\end{algorithm}

\begin{figure}[ht]
	\centering
	\includegraphics[width=0.2\textwidth]{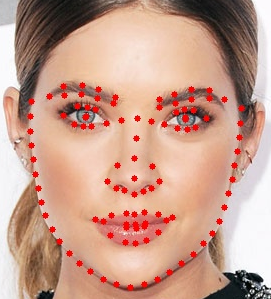}
	\caption{106 landmarks.}
	\label{fig:landmarks}
\vspace{-0.1in}
\end{figure}

After obtaining face rectangles, $\{F_j\}$, we exploit face alignment algorithm to estimate the location of 106 facial landmarks~\cite{zhuunconstrained} on every rectangle. The locations of these landmarks are shown in Figure \ref{fig:landmarks}. 

Further, we use alignment results to regularize every rectangle.
Particularly, we formalize the landmarks on $j$-th face as $L_j=(l_1,l_2,\cdots,l_{106})$, where $l_i$ denotes $(x_i, y_i)^\top$, the coordinates of $i$-th landmark. And, we calculate the transformation matrix $T_j$ by:

\begin{equation}
\notag
\begin{array}{c@{\quad}l@{\quad}l}
& T_j &= \mathop{\argmin}_{T} || T \tilde{L}_{j} - L_{mean} || ^ 2 \\
\mbox{where} &

L_{mean} &= \frac{\sum_{j} {L_j}}{\sum_{j} {1}} \\

&\tilde{L}_{j} &= 
\left[
\begin{matrix}
x_1 & x_2 & \cdots & x_{106} \\
y_1 & y_2 & \cdots & y_{106} \\
1 & 1 & \cdots & 1 \\
\end{matrix}
\right]
\end{array}
\end{equation}
where $T$ is a $3 \times 3$ matrix contains rotation and shifting coefficients. We select the best $T$ as $T_j$ that minimizes the $L_2$ distance between the regularization target $T_{mean}$ and $\tilde{L}_{j}$, the homogeneous matrix of the coordinate matrix.
After that, we regularize the $j$-th frame by applying the transformation matrix $T_j$ to every pair of coordinates and extract the centering 1280x720 rectangle containing the face. For the sake of simplicity, we use "frame" to represent these regularized frames containing only the face~\footnote{Since we just implemented those existing algorithms on face tracking, detection and alignment, we will not provide further details about them and interested readers can refer to original papers.}.

\subsubsection{\textbf{Timing Verification}}
\label{sec:timeValidation}

Our timing verification is built on the nature of how camera and screen work. Basically, both of them follow the same scheme: refreshing pixel by pixel. Detailedly, after finishing refreshing one line or column, they move to the beginning of next line or column and perform the scanning repeatedly. We can simply suppose an image is displayed on screen line by line and captured by camera column by column, ignoring the time gap between refreshing adjacent pixels within one line or column that is much smaller than the time needed to jump to the next line or column. In other words, as to update any specific line on the screen, it has to wait for a complete frame cycle until all other lines have been scanned. Similarly, when a camera is capturing an image, it also has to wait for a frame cycle to refresh a certain column.

\begin{figure}[htb]
	\centering
  \begin{subfigure}{0.25\textwidth}
		\includegraphics[width=\textwidth]{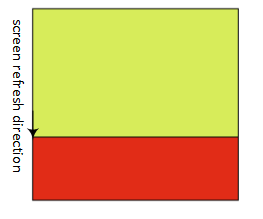}
    \caption{screen refreshing}
    \label{fig:d_screen}
  \end{subfigure}
  \begin{subfigure}{0.22\textwidth}
		\includegraphics[width=\textwidth]{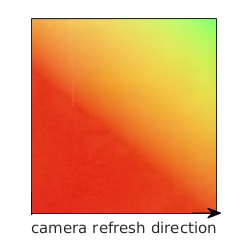}
    \caption{camera refreshing}
    \label{fig:d_camera}
  \end{subfigure}
  
  \caption{Working schemes of screen and camera.}
  \label{fig:d_ROI}
\end{figure}

One example is given in Fig~\ref{fig:d_ROI} to better explain the interesting phenomenon that is leveraged for our timing verification. Fig~\ref{fig:d_screen} shows a screen that is just changing the displaying color from Red to Green. Since it is scanning horizontally from top to bottom, the upper part is now updated to Green but the lower part is still previous color Red. The captured image of a camera with column scanning pattern from left to right is shown in Fig~\ref{fig:d_camera}, which shows an obvious color gradient from Red to Green~\footnote{Similar but a little bit different color patterns can also be observed on cameras with row scanning mode. Column scanning mode is used here it is easier to understand.}.

\begin{figure}[htb]
	\centering
  \begin{subfigure}{0.25\textwidth}
		\includegraphics[width=\textwidth]{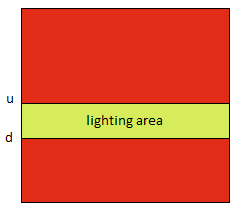}
		\caption{lighting challenge}
    \label{fig:lightingArea}
  \end{subfigure}
  \begin{subfigure}{0.19\textwidth}
		\includegraphics[width=\textwidth]{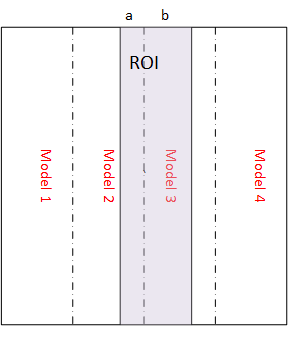}
		\caption{captured frame}
    \label{fig:calculateROI}
  \end{subfigure}
  \caption{Example of lighting area and calculation of ROI.}
  \label{fig:infer}
\end{figure}

To transform this unique feature into a strong security guarantee, the appropriate challenges must be constructed and verified to ensure the consistency of responses. In practice, we construct two types of challenge to be presented on front-end screen: one is the background challenge displaying a single color, and the other is the lighting challenge displaying a belt of different color on the background color. The belt of the different color from background is called the \textit{lighting area}, and one example is shown in Fig~\ref{fig:lightingArea}, where the background color is Red while the lighting area is Green.

To verify the consistency in responses, we defined another concept called \textit{Region of Interest} (ROI), which is the region that the camera is scanning when the front-end screen is displaying the lighting area. The location of ROI is calculated as followings: 
\begin{itemize}

\item Calculate $t_u$, the start time to show the lighting area.
\begin{equation}
  t_u = t_{begin} + \frac{u}{rows} * t_{frame}
\end{equation}
where $u$ is the upper bound of lighting area, $rows$ is the number of rows contained in one frame, $t_{begin}$ is the start time to show the current frame, and $t_{frame}$ is the during time of one frame.

\item Find the captured frame which recording period covers $t_u$. Say the $k$-th captured frame.

\item Calculate the shift, $l$, against the first column of $k$-th captured frame.
\begin{equation}
  l = cols * \frac{t_u - ct_{k} }{ ct_{frame} }
\end{equation}
where $cols$ is the number of columns contained in one captured frame, $ct_k$ is the start time to exposure the first column of $k$-th capture frame, and $ct_{frame}$ is the exposure time of one captured frame.
\end{itemize}

\begin{figure}[htb]
	\centering
  \begin{subfigure}{0.2\textwidth}
		\includegraphics[width=\textwidth]{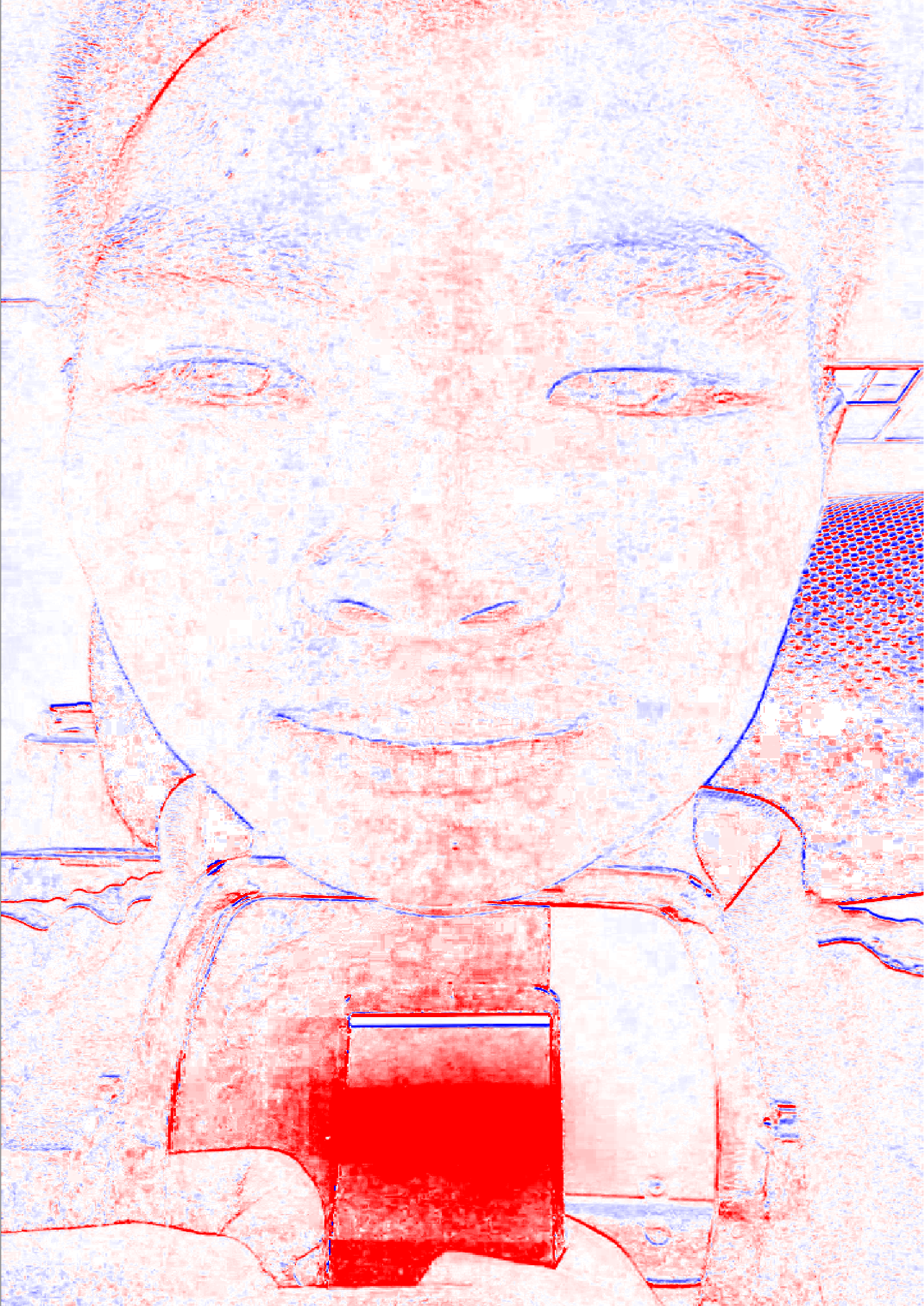}
    	\caption{light middle area}
  \end{subfigure}
  \begin{subfigure}{0.2\textwidth}
		\includegraphics[width=\textwidth]{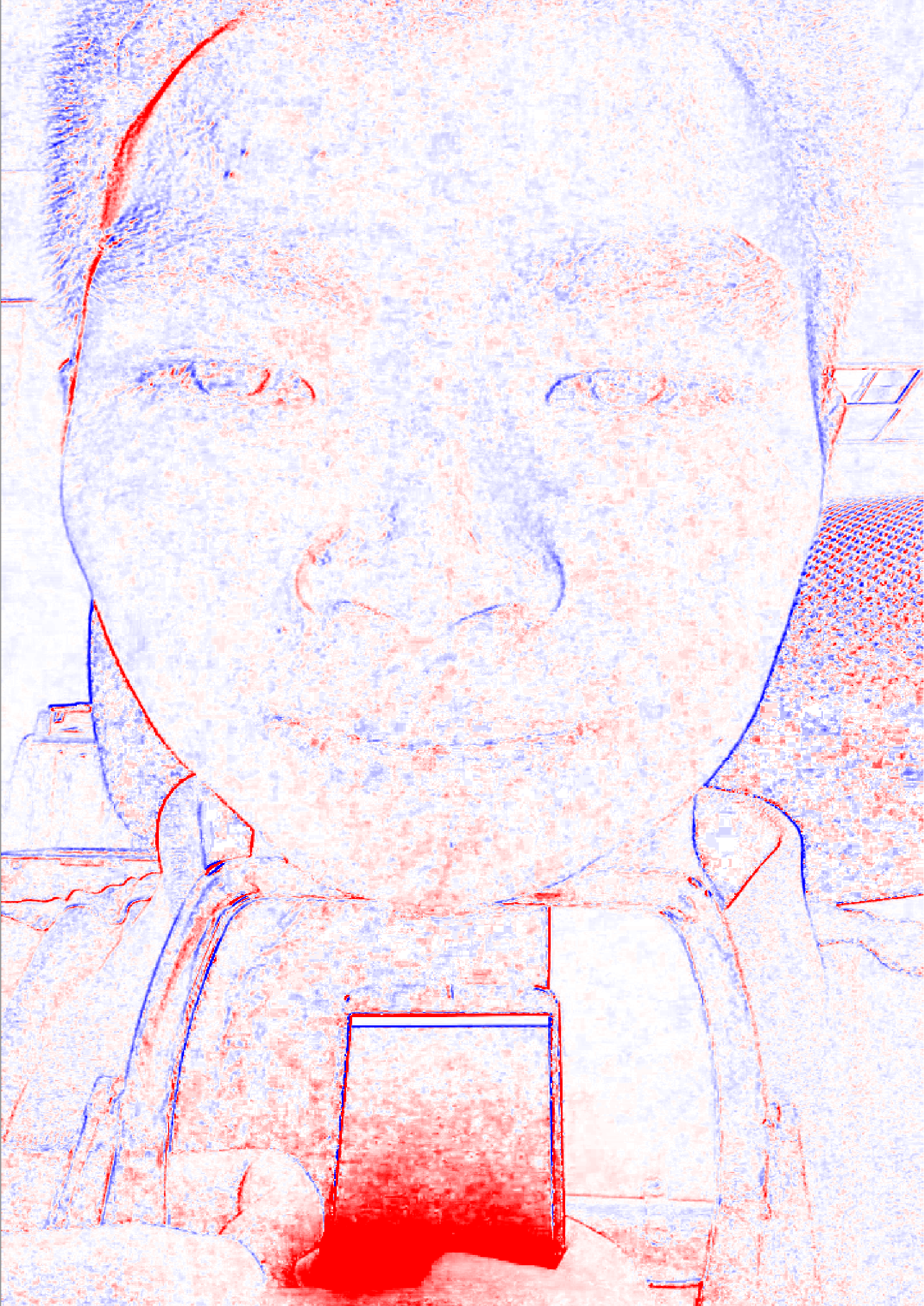}
    	\caption{light bottom area}
  \end{subfigure}
  
  \caption{Effect of lighting area. In the bottom of both pictures, these are mirrors showing the location of corresponding lighting area.}
 
  \label{fig:light}
\end{figure}

After finding the location of ROI, we distill it by applying Eq.(\ref{fa:time}) on every pixel between the response of lighting challenge and background challenge. Two applied results are demonstrated on Fig~\ref{fig:light}. Now, the consistence can be verified. We check whether the lighting area can be correctly inferred from the distilled ROI. If it cannot, the delay exists and this response is counterfeit. 

To infer the lighting area, we build 4 linear regression models handling different part of captured frame (Fig~\ref{fig:calculateROI}). Each model is fed a vector, the average vector reduced from corresponding part of ROI, and estimates the location of $\frac{u+d}{2}$ independently. Next we gather estimated results according to the size of each part. An example is shown on Fig~\ref{fig:calculateROI} where the ROI is separated into 2 parts: the left part contains $a$ columns and the right part contains $b$ columns. The gathered result, $\hat{y}$, is calculated as following. 
\begin{equation}
\hat{y} = \frac{a \times m_2 + b \times m_3}{a+b}
\end{equation}
where $m_2$ and $m_3$ denote the estimated result made by model 2 and model 3 respectively. 

The final criteria of consistence is accumulated from $\hat{y_i}$, the gathered result of $i$-th captured frame, as following:
\begin{equation}
\begin{array}{c@{\quad}l@{\quad}l}
d_i = \hat{y_i} - \frac{u_i + d_i}{2} \\
mean_d = \frac{\sum_{i=1}^{n} d_i}{n} \\
std_d^2 = \frac{\sum_{i=1}^{n} (d_i-mean_d ) ^2}{n-1}
\end{array}
\end{equation}
We finally check whether $mean_d \times std_d$ is smaller than $exp(Th)$, where $Th$ is a predefined threshold.

Note that legitimate responses are consistent with our challenges and will produce both small $mean_d$ and $std_d$. Adversarial responses will be detected by checking our final criteria. An additional demo was illustrated on Fig~\ref{fig:light} to explain visually how the lighting area affects the captured frame.

\subsubsection{\textbf{Face Verification}}
\label{faceValidation}

\begin{figure*}
	\centering
  \begin{subfigure}{0.25\textwidth}
		\includegraphics[width=\textwidth]{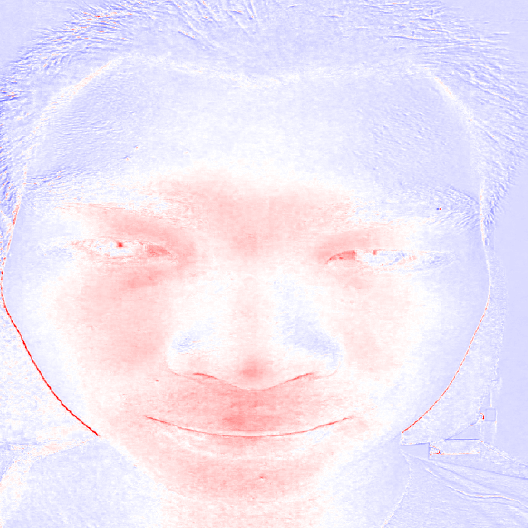}
    \caption{}
    \label{fig:R_example:a}
  \end{subfigure}
  \begin{subfigure}{0.25\textwidth}
		\includegraphics[width=\textwidth]{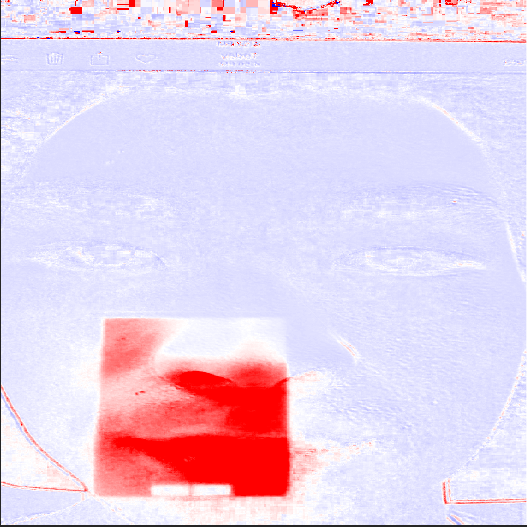}
    \caption{}
    \label{fig:R_example:b}
  \end{subfigure}
  \begin{subfigure}{0.25\textwidth}
		\includegraphics[width=\textwidth]{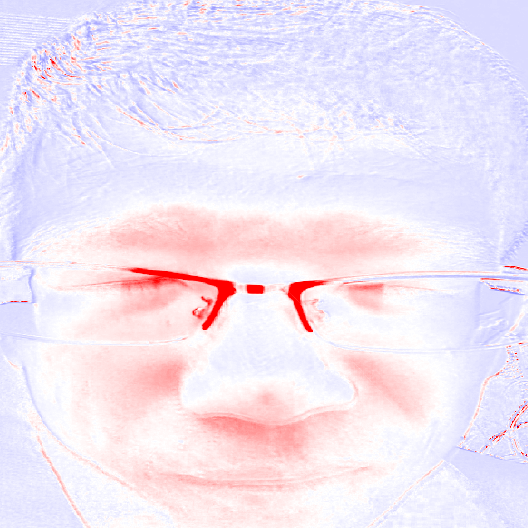}
    \caption{}
    \label{fig:R_example:c}
  \end{subfigure}
  \\
  \begin{subfigure}{0.25\textwidth}
		\includegraphics[width=\textwidth]{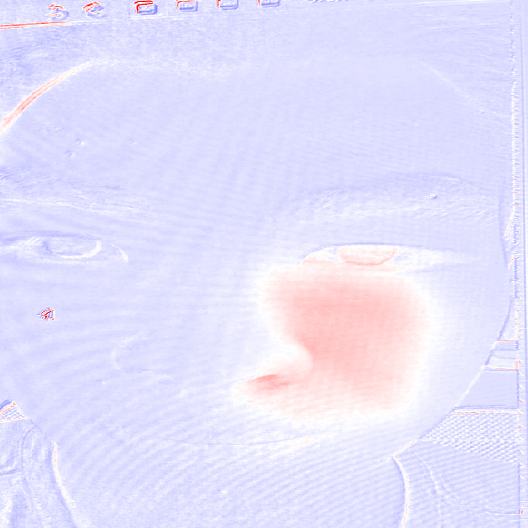}
    \caption{}
    \label{fig:R_example:d}
  \end{subfigure}
  \begin{subfigure}{0.25\textwidth}
		\includegraphics[width=\textwidth]{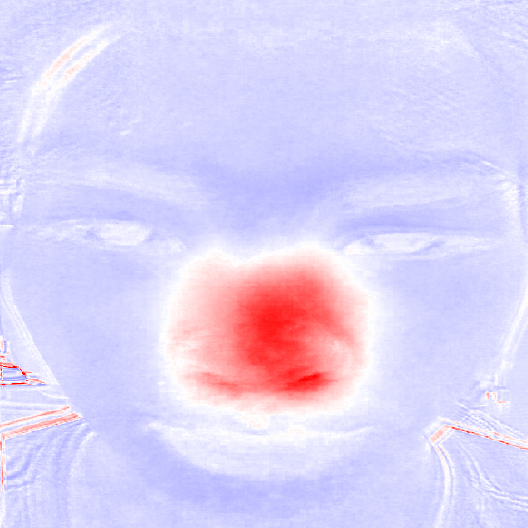}
    \caption{}
    \label{fig:R_example:e}
  \end{subfigure}
   \begin{subfigure}{0.25\textwidth}
		\includegraphics[width=\textwidth]{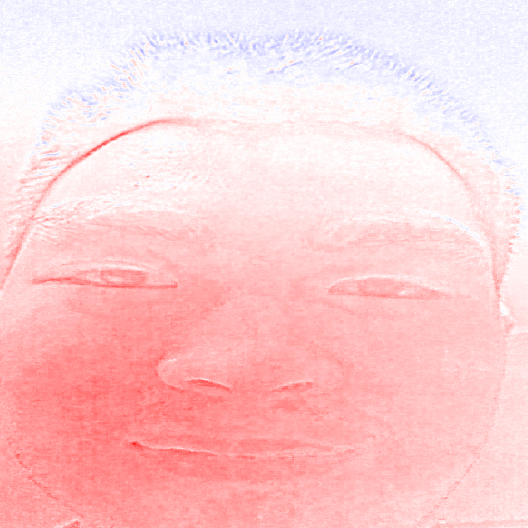}
    \caption{}
    \label{fig:R_example:f}
  \end{subfigure}

	\caption{Examples of midterm results. (a) and (c) are captured from real human faces, (b) is captured from an iPad's screen, (d) and (e) are captured from a LCD monitor, and (f) is captured from a paper.}
	\label{fig:R_example}
\vspace{-0.2in}
\end{figure*}

After preprocessing, we get a sequence frames with vibration removed, size unified and color synchronized.
Further, we use Eq.(\ref{fa:geometry}) to generate the midterm result from the responses of a background challenge: First, we randomly choose a pixel on the face as the anchor point; then, we divide all the pixels by the value of that anchor point. Some midterm results are shown on Fig~\ref{fig:R_example}.

Without any difficulty, we can quickly differentiate results of real human faces from fake ones. This is because real human faces have uneven geometry and textures, while other materials, like monitor, paper or iPad's screen, do not have. Based on this observation we developed our face verification techniques, as described below.

\begin{figure}
	\centering
  \begin{subfigure}{0.23\textwidth}
		\includegraphics[width=\textwidth]{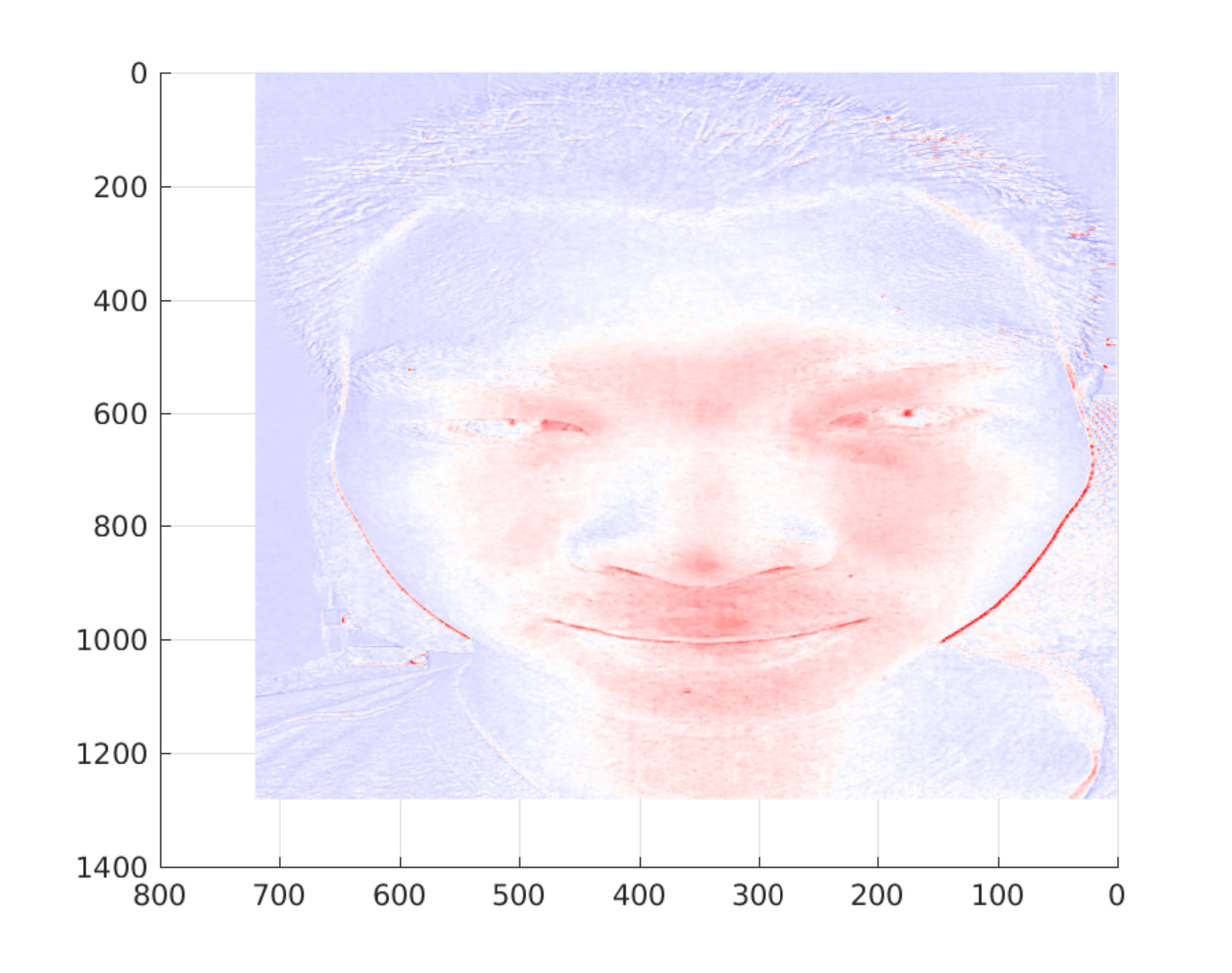}
    \caption{midterm result}
    \label{fig:reduce_1}
  \end{subfigure}
  \begin{subfigure}{0.23\textwidth}
		\includegraphics[width=\textwidth]{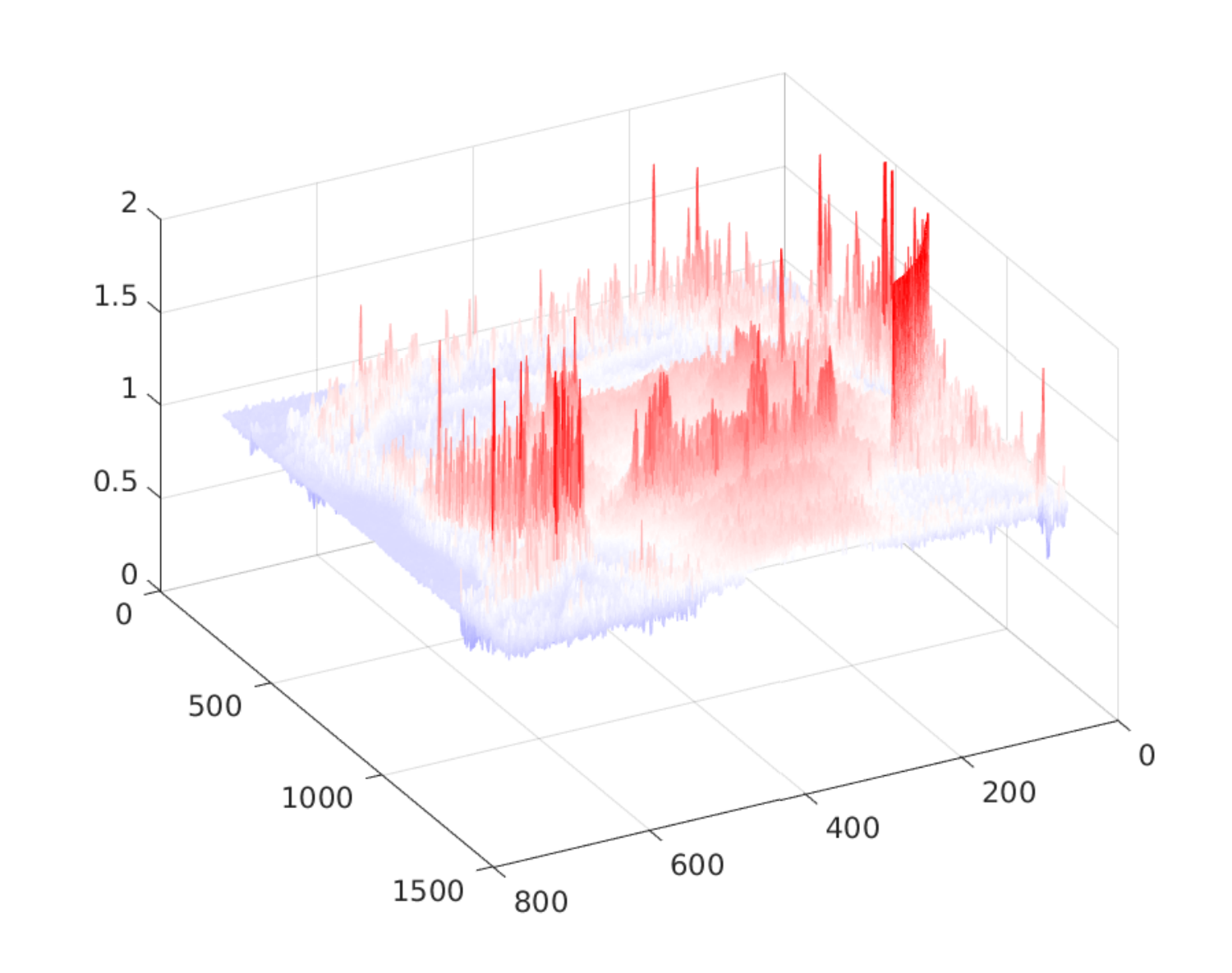}
    \caption{midterm result}
    \label{fig:reduce_2}
  \end{subfigure}
  \\
  \begin{subfigure}{0.23\textwidth}
		\includegraphics[width=\textwidth]{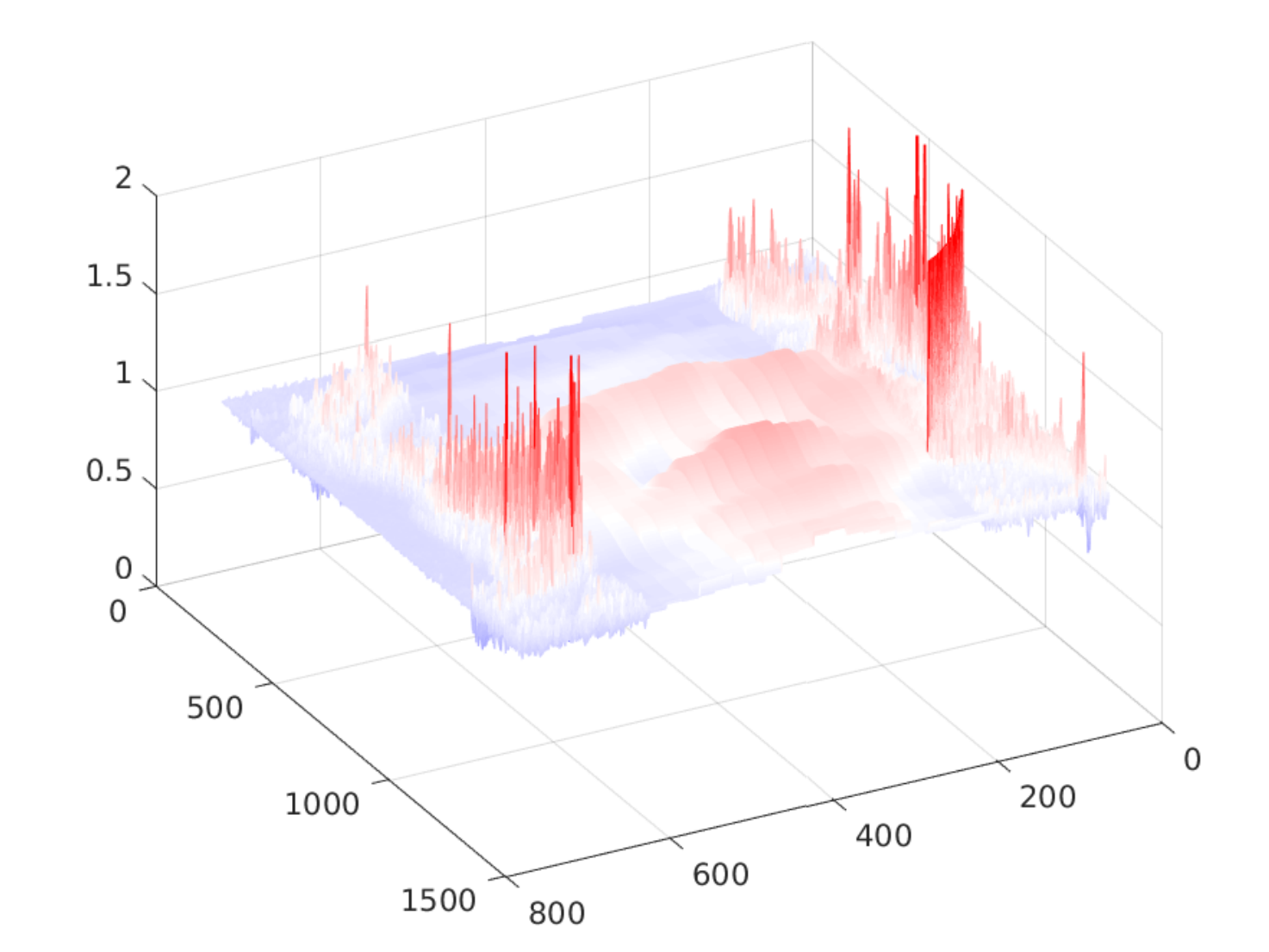}
    \caption{abstract result}
    \label{fig:poly}
  \end{subfigure}
  \begin{subfigure}{0.23\textwidth}
		\includegraphics[width=\textwidth]{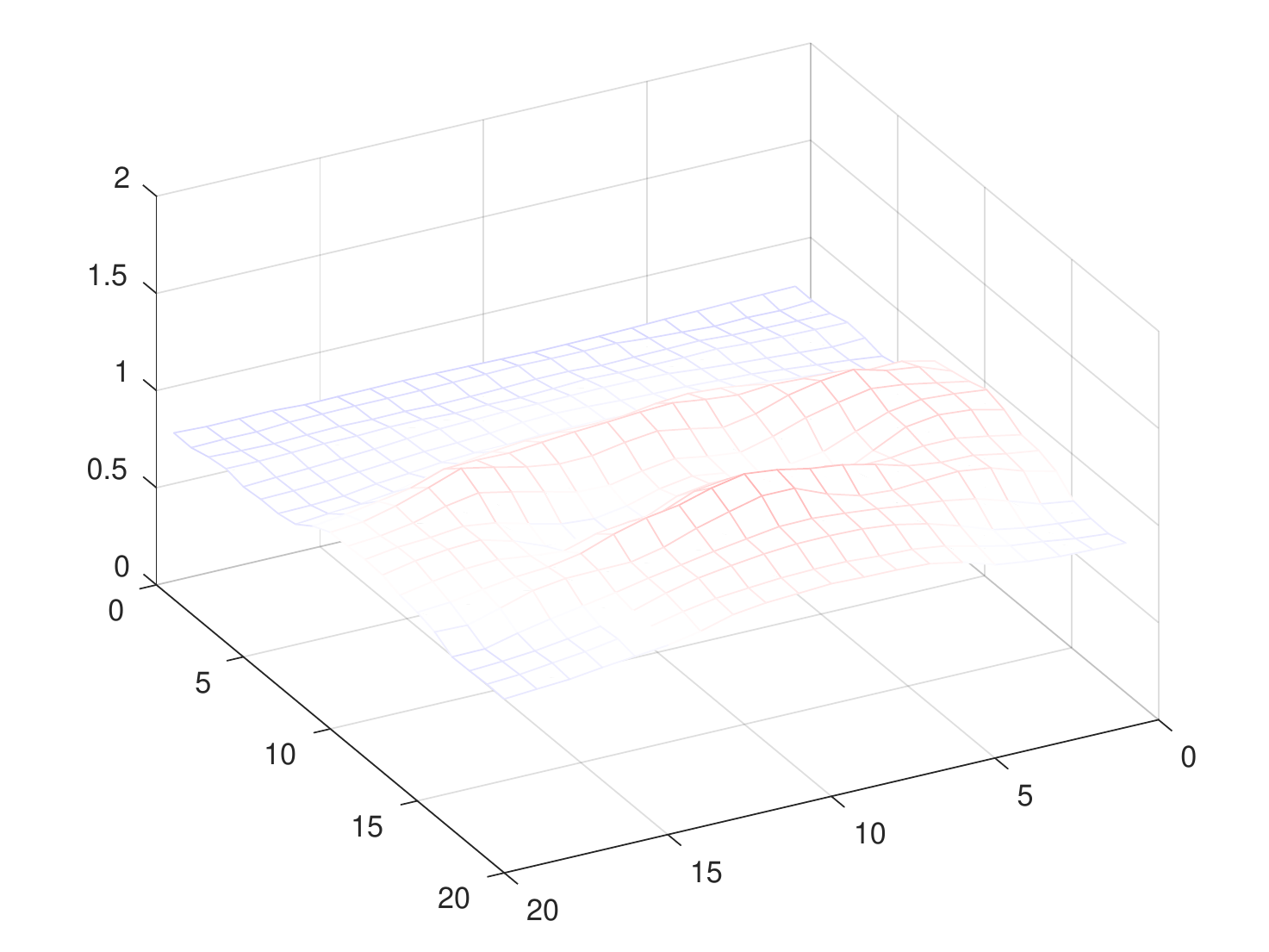}
    \caption{resize result}
    \label{fig:twenty}
  \end{subfigure}

	\caption{Face verification.}
	\label{fig:reduce}
\end{figure}

\begin{itemize}
\item Step 1: abstract. We vertically divide the face into 20 regions. In every region, we further reduce the image to a vector by taking the average value. Next, we smooth every vector by performing polynomial fitting of 20 degrees with minimal 2-norm deviation. After that, we will derive images like Fig~\ref{fig:poly}.

\item Step 2: resize. We pick out facial region and resize it to a 20x20 image by bicubic interpolation. An example is shown on Fig~\ref{fig:twenty}.

\item Step 3: verify. We feed the resized image to a well-trained neural network, and the decision will be made then.
\end{itemize}

The neural network we used contains 3 convolution layers with a pyramid structure, which effectiveness was sufficiently proved in Cifar-10, a dataset used to train the neural network to classify 10 different objects. In Table~\ref{tb:nn}, we show the architecture of our network and the parameters of every layer.

\begin{table}[tbh]
  \centering
    \caption{Architecture of Neural Network.}
	\begin{tabular}{|c|c|c|c|}
	\hline
	input size & layer type & stride & padding size\\
	\hline
	20x20x3 & conv 5x5 & 1 & 0 \\
	\hline
	16x16x16 & conv 3x3 & 1 & 1 \\
	\hline
	16x16x16 & pool 2x2 & 1 & 0 \\
	\hline
	8x8x16 & conv 3x3 & 1 & 1 \\
	\hline
	8x8x32 & pool 2x2 & 1 & 0 \\
	\hline
	1x512 & inner product & 0 & 0 \\
	\hline
	\end{tabular}
	
	\label{tb:nn}
\vspace{-0.1in}
\end{table}

\section{Security Analysis}
\label{sec:securityAnalysis}

In this section, we present the security analysis of Face Flashing. First, we 
abstract the mechanisms behind Face Flashing as a challenge-response protocol.
Second, we analyze the security of two main parts in our protocol: timing verifications and face verification. Finally, we demonstrate how Face Flashing defeats three typical advanced attacks.

It is certain that Face Flashing can defeat static attacks, as the expression detector, one component of our system, is sufficient to defeat them. Specifically, static materials cannot make expressions according to our instructions in time (e.g., 1 second) and attacks using them will be failed by expression detector. Besides, we conduct a series of experiments in Section~\ref{sec:evaluation} to demonstrate that the expression detector can be correctly integrated with our other verifications. Therefore, the main task of our security analysis is to show that Face Flashing can defeat dynamic attacks.


\subsection{A Challenge-Response Protocol}
\label{sec:securityModel}

Face Flashing is a challenge-response protocol whose security
guarantees are built upon three elements: unpredictable random challenges,
hardly forged responses, and the effective response verifications.

\bheading{The Challenges}. 
Our challenge is a sequence of carefully-crafted images that are generated at random. Since the front-end devices are assumed to be well protected, adversaries could not learn the random values. Besides, a verification session consists of tens of challenges. Even if the adversary can respond a right challenge by chance, it is unlikely for him to respond to a sequence of challenges correctly.

\bheading{The Responses}. 
There are two important requirements for the responses: 
First, the response must be easily generated by the legitimate users, otherwise it may lead to usability problems or even undermine the security guarantee (e.g., if adversaries can generate fake responses faster than legitimate users).
Secondly, the responses should include essential characteristics of the user, which are hard to be forged.

Face Flashing satisfies both requirements. The response is the reflected light from the human face, and the user needs to do nothing besides placing her face against the camera. More importantly, such responses, in principle, are generated at the speed of light, which is faster than any computational process. Besides, the response carries unique characteristics of the subject, such as the reflectance features of her face and uneven geometry shapes, which are physical characteristics of human faces that are inherently different from other media, e.g., screens (major sources of security threats). 

\bheading{Response Verification.} 
We use an in-depth defense strategy to verify the responses and detect possible attacks.
\begin{itemize}
\item First, timing verification is used to prevent forged responses (including replay attacks). 
 
\item Second, face verification is used to check if the subject under authentication has a specific shape similar to a real human face.

\item Third, this face-like object must be regarded as the same person with the victim by the face recognition module (orthogonal to liveness detection).

\end{itemize}

Considering the pre-excluded static object, it is very hard for adversaries to fabricate such a thing satisfying 3 rules above simultaneously. In general, Face Flashing builds a high bar in front of adversaries who want to impersonate the victim.

\subsection{Security of Timing Verification}
\label{sec:timeAnalysis}

The goal of the timing verification is to detect the delay in the response time caused by adversaries. Before further analysis, we emphasize two points should be considered.

\begin{itemize}
\item First, according to the design of modern screens, the adversary cannot update the picture that is being displayed on the screen at the arbitrary time. In other words, the adversary cannot interrupt the screen and let it show the latest generated response before the start of next refreshing period.

\item Second, the camera is an integral device which accumulates the light during his exposure period. And, at any time, within an initialized camera, there always exists some optical sensors are collecting the light.
\end{itemize}

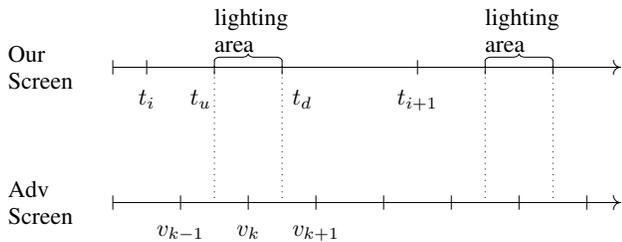
\begin{figure}
\scalebox{0.9}{
    \begin{tikzpicture}
        \draw (0,0) -- (7.5,0);
        \draw (0,-2) -- (7.5,-2);

        \foreach \x in {0,0.5,1.5,2.5,4.5, 5.5, 6.5}
          \draw (\x cm,3pt) -- (\x cm,-3pt);
          
        \foreach \x in {0,1,2,3,4,5,6,7}
          \draw (\x cm,-2cm + 3pt) -- (\x cm, -2cm -3pt);


        \foreach \y in {0,-2}
            \draw[join=round,cap=round] (7.5cm - 3pt,\y cm + 3pt) arc (180:270:3pt) arc (90:180:3pt);

        \draw[dotted] (1.5, 0cm + 3pt) -- (1.5, -2cm);  
        \draw[dotted] (2.5, 0cm + 3pt) -- (2.5, -2cm);   
        \draw[dotted] (5.5, 0cm + 3pt) -- (5.5, -2cm);  
        \draw[dotted] (6.5, 0cm + 3pt) -- (6.5, -2cm);   
        
        \draw[snake=brace] (1.5, 0cm + 3pt) -- (2.5,  0cm +3pt);
        \draw[snake=brace] (5.5, 0cm + 3pt) -- (6.5,  0cm +3pt);


        \draw (0,0) node[text width=1cm,left=12pt] {Our Screen};
        \draw (0,-2) node[text width=1cm,left=12pt] {Adv Screen};
        \draw (2,0) node[text width=1cm, above=3pt] {lighting area};
        \draw (6,0) node[text width=1cm, above=3pt] {lighting area};
        \draw (0.5,0) node[below=6pt]{$t_{i}$};
        \draw (4.5,0) node[below=6pt]{$t_{i+1}$};
        \draw (1.3,0) node[below=6pt]{$t_{u}$};
        \draw (2.8,0) node[below=6pt]{$t_{d}$};
        \draw (1,-2) node[below=6pt]{$v_{k-1}$};
        \draw (2,-2) node[below=6pt]{$v_{k}$};
        \draw (3,-2) node[below=6pt]{$v_{k+1}$};
        
    \end{tikzpicture}
}
  \caption{Security analysis on time.}
  \label{fig:timeAnalysis}
 \vspace{-0.1in}
\end{figure}

For sake of clarity, we assume the front-end devices contain a 60-fps camera and 60-Hz screen. On the other side, the adversary has a more powerful camera with 240-fps and screen with 240-Hz. Under these settings, we construct a typical scenario to analyze our security, which time lines are shown on Fig~\ref{fig:timeAnalysis}.

In this scenario, the screen of the front-end device is displaying the $i$-th challenge, 
and the adversary aims to forge the response to this challenge. The adversary may instantly learn the location of lighting area of the challenge after $t_u$. While she cannot present the forged response on her screen until $v_k$, due to the nature of how the screen works. Hence, there is a gap between $t_u$ and $v_k$. Recalling our method described in Section~\ref{sec:timeValidation}, during the gap, some columns in the ROI have already completed the refreshing process. In other words, these columns' image will not be affected by the forged response displaying on the adversary's screen during $v_k$ to $v_{k+1}$. We name this phenomenon as \textit{delay}.

When \textit{delay} happens, our camera will get an undesired response inducing four linear regression models to do deviated estimation about the location of lighting area. Besides, the standard deviation of these estimations will increase, for two reasons:
\begin{itemize}
\item The adversary's screen can hardly be synchronized with our screen. Particularly, it is different even the length of adjacent refreshing periods. Hence, the $delay$ is unstable, so as the estimations.
\item The precision of forging will be affected by the internal error of adversaries' measurement about time. This imprecision will be amplified again by our camera, which fluctuates the estimations. 
\end{itemize}

In other words, if the adversary reduces $mean_d$ by displaying the carefully-forged response, she will simultaneously increase $std_d$. On the other hand, if the adversary does nothing to reduce $std_d$, she will significantly enlarge $mean_d$. While for a benign user, the \textit{delay} will not happen, the discordance between our camera and screen can be solved by checking the timestamps afterward, and both the accumulated $mean_d$ and $std_d$ will be small, according to our verification algorithm.

In summary, we detect the delay by estimating the deviation. And the effectiveness of our algorithm provides a strong security guarantee on the timing verification.

\subsection{Security of Face Verification}
\label{faceAnalysis}

Our face verification abstracts the intrinsic information of shape through a series of purification. And we feed this information to a well-designed neural network.

If the adversary aims to bypass the face verification, there are two conundrums that need to be resolved. First, the adversary needs to conceal the specular reflection of the plain screen. Particularly, during the authentication procedure, we require the user to hold the phone so their face can occupy the entire screen. The distance, as we measured, is about 20-cm. In this short distance, the specular reflection is severe. In Fig~\ref{fig:R_example:b}, we demonstrate the result captured from a screen without any covering sheet. Even covered by a scrub film (Fig~\ref{fig:R_example:d}), the screen's specular reflection is still intense.

Second, the forged object must have similar geometry shape with human faces. More precisely, its abstract result should like a transpose of ``H'' (Fig~\ref{fig:poly}). And this stereo object needs to make expression according to our instructions. Even if the adversary can achieve these, there is no promise they can deceive our strong neural network modal every time. And there is no chance for the adversary to generate a response with low quality. The high recall of our model will be demonstrated in next section. 

The above two conundrums provide the security guarantee on face verification.

\subsection{Security against typical attacks}
\label{sec:secureAttack}

Obviously, Face Flashing can defeat traditional attacks like photo-based attacks. 
Here we discuss its defenses against three typical advanced attacks: 

\bheading{Offline Attacks.} An offline attack is to record responses of previous authentications, and replay them to attack the current authentication. However, this attack is impossible to fool our protocol. First, the hitting possibility is small, as we require responses match all the challenges. Concretely, if we use 8 different colors and present 10 lighting challenges, the hitting possibility will be less than $10^{-9}$. Second, even if adversaries have successfully guessed the correct challenge sequence, displaying responses legitimately is difficult. Because displaying on screens will produce the intensely specular reflection that is easily detectable, and displayed by projecting responses onto a forged object leads to high $std_d$ that also can be detected, as adversaries cannot precisely predict the length of every refreshing period of the screen. 

\bheading{MFF Attacks.} An MFF attack is to forge the response by merging victim's facial information and the currently received challenge. However, this attack is also useless, because it is hard to deceive our timing and face verifications simultaneously. First, to deceive our face verification needs forging high-quality responses which is difficult and time-consuming. Particularly, high-quality forgery requires reconstructing the 3D model of victim's face and simulating the reflection process. Second, to deceive our timing verification needs to complete the above forgery quickly. Actually, the available time is $\frac{1}{240}/2$ second for attacking a 60 Hz screen (Section~\ref{sec:eval:time}). Third, even if adversaries can quickly produce a perfectly forged response, displaying the response is not allowed (see the preceding paragraph).  

\bheading{3D-Mask Attacks.} A 3D-mask attack is to wear a 3D mask to impersonate the victim. However, this attack is impractical. First, this attack needs to build an accurate mask that can fool our face recognition module, which is difficult\footnote{Even though there is an existing study implying it is possible~\cite{raghavendra2014robust}, performing it in real is not easy.}. Second, the legitimate mask is hard to be 3D printed. As the printed mask needs to have the similar reflectance of human skin and be so flexible that adversaries can wear it to make instructed expressions. While, the available 3D printed materials are non-flexible under the requirement of Fused Deposition Modeling (FDM), the prevalent 3D print technology. Besides, the smallest diameter of available nozzles is 0.35mm that will produce coarse surfaces, and coarse surfaces can be distinguished from human skin. 

In sum, Face Flashing is powerful to defeat advanced attacks, especially attacks similar to the ones mentioned above.

\section{Implementation and Evaluation}
\label{sec:evaluation}

In this section, we introduce the source of our collected data at the beginning, then present implementations and evaluations of timing and face verifications, followed by the evaluation on robustness. Finally, we give the computational and storage cost when deploying our system on a smartphone and the back-end server.

\subsection{Data Collection}
We have invited 174 participants including Asian, European and African. Among all participants, there are 111 males and 63 females with ages ranging from 17 to 52. 
During the experiment, participants were asked to hold a mobile phone facing to their face and make expressions such as smiling, blinking or moving head slightly. A button was located at the bottom of the screen so that participants can click it to start (and stop) the authentication/liveness detection process. When started, the phone performs our challenge-response protocol and records a video with its front camera. And, once started, that button will be disabled for three seconds to ensure that every captured video contains at least 90 frames.

In total, we collect 2274 raw videos under six different settings (elaborated in Section~\ref{sec:val:robust}). In each scenario, we randomly select 50 videos to form the testing data set, and all other videos then belong to the training data set.

\subsection{Timing Verification}
\label{sec:eval:time}

In our implementation of timing verification, we set the height of lighting area in every lighting challenge to a constant, i.e., $u-d = 1/4$, where the height of the whole screen is $1$. And we use an open source library, \textit{LIBLINEAR}~\cite{REF08a},  to do the regression with $L2$-loss and $L2$-regularization, where the $Th$ is set to $-5$.

\begin{figure}
	\centering
  \begin{subfigure}{0.22\textwidth}
		\includegraphics[width=\textwidth]{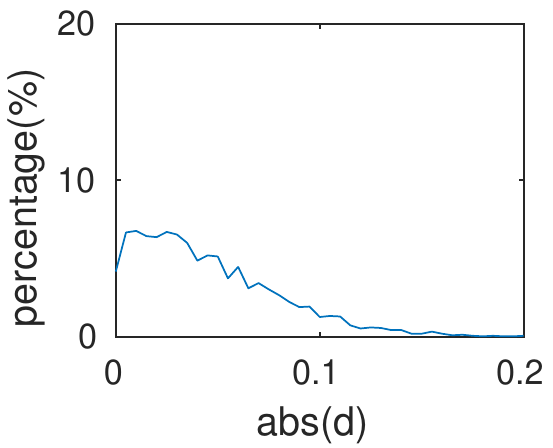}
    \caption{mean=0.046, std=0.035}

  \end{subfigure}
  \begin{subfigure}{0.22\textwidth}
		\includegraphics[width=\textwidth]{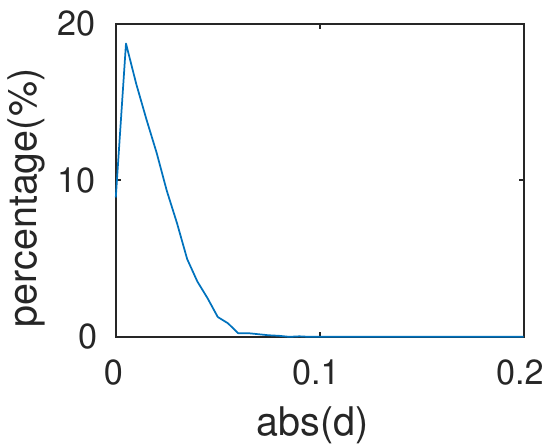}
    \caption{mean=0.012, std=0.013}

  \end{subfigure}
  \begin{subfigure}{0.22\textwidth}
		\includegraphics[width=\textwidth]{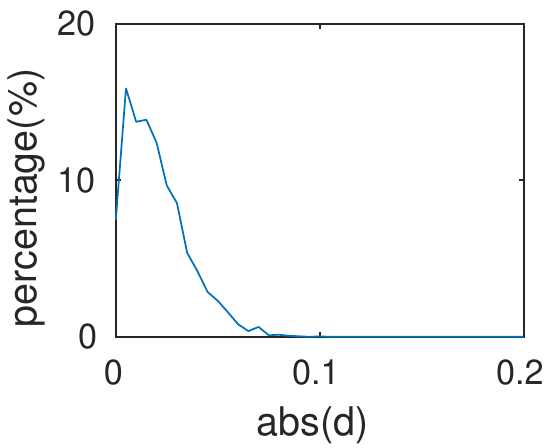}
    \caption{mean=0.020, std=0.015}

  \end{subfigure}
  \begin{subfigure}{0.22\textwidth}
		\includegraphics[width=\textwidth]{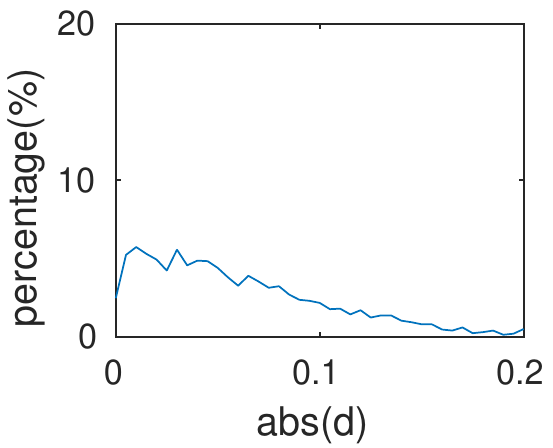}
    \caption{mean=0.060, std=0.045}

  \end{subfigure}

	\caption{Performance of 4 regression models. (a)-(d) shows performance of model 1-4 respectively.}
	\label{fig:regression}
\vspace{-0.1in}
\end{figure}

We trained four regression models on the training set mentioned above, and their performances over the testing data set are shown on Fig~\ref{fig:regression}. 
It shows that performances of model 1 and 4 are relatively poor which is reasonable in fact, because both models handle two challenging areas (refer to Fig~\ref{fig:infer}) where the responses are weak and the keen edges also impair the results.

\begin{figure}
	\centering
  \begin{subfigure}{0.18\textwidth}
		\includegraphics[width=\textwidth]{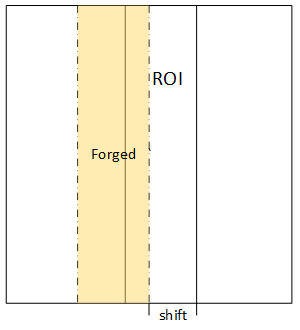}
	\caption{}
    \label{fig:attack:1}
  \end{subfigure}
  \begin{subfigure}{0.22\textwidth}
		\includegraphics[width=\textwidth]{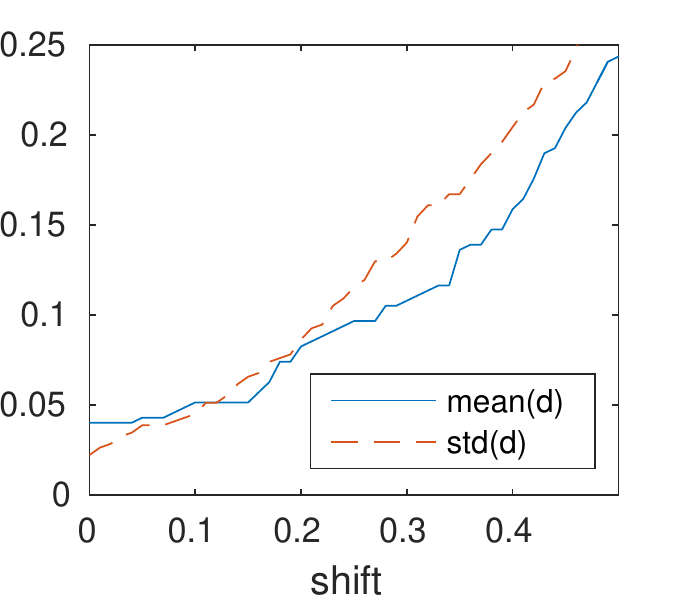}
	\caption{}
    \label{fig:attack:2}
  \end{subfigure}
	\caption{Attack simulation}
	\label{fig:attack}
\vspace{-0.1in}
\end{figure}

To evaluate its capability on defending against attacks, we feed forged areas (see Fig~\ref{fig:attack:1}) to these regression models, and observe the results. It turns out that when enlarging the $shift$ between real ROI and forged area, the estimation deviation increases. In Fig~\ref{fig:attack:2}, we illustrated the relationship between estimated $mean_d$ and $std_d$ under different values of $shift$, while regularizing the width of ROI as 1.
The figure shows that when $shift$ is less than $0.1$, the estimation error of $mean_d$ and $std_d$ is very small. But  when the shift is $0.5$, the estimation error is around $1/4$. In other words, when increasing $shift$ to the half of ROI's width, the estimated deviation could be larger than the height of the lighting area, which states that adversary's opportunity window (i.e., $shift$) for a successful attack is pretty small, and our method can reliably detect such attacks. Concretely, the acceptable delay for a benign response is less than $\frac{1}{240}/2$ second for a 60 Hz screen.

\begin{figure}
	\centering
  \begin{subfigure}{0.33\textwidth}
		\includegraphics[width=\textwidth]{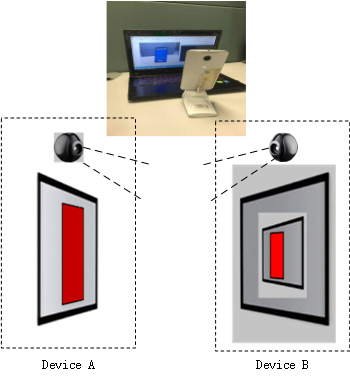}
	\caption{scenario}
    \label{fig:timeAttack:1}
  \end{subfigure}
  
  \begin{subfigure}{0.33\textwidth}
		\includegraphics[width=\textwidth]{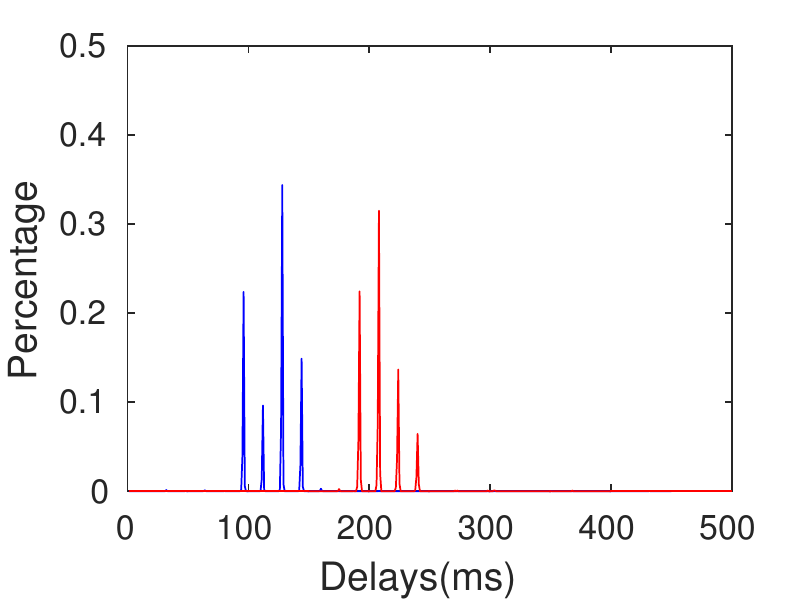}
	\caption{results}
    \label{fig:timeAttack:2}
  \end{subfigure}
  
	\caption{Primitive attack.}
	\label{fig:timeAttack}
\vspace{-0.3in}
\end{figure}

Further, we investigated the delays under a real-world setting (shown in Fig~\ref{fig:timeAttack}). In this experiment, we used two devices: A is the authenticator (a Nexus 6 smartphone in this example), and B is the attacker (a laptop that will reproduce the color displayed on smartphone by simply showing the video captured by its front camera). 
When the experiment begins, the smartphone starts to flash with random colors, and record whatever is displayed on laptop screen at the same time, then calculate the delay needed by attackers to reproduce the same color. The same procedure will be repeated to calculate the delays by replacing the laptop with a mirror.

Fig~\ref{fig:timeAttack:2} shows the results where the blue bars are mirror's delays while the red bars are the laptop's delays. The difference between the delays means that if adversaries had used devices other than mirrors to reproduce the reflected colors (i.e. responses), there should be significant delays. This is actually one of our major technical contribution to use light reflections instead of human expressions and/or actions as the responses to given challenges, and it can give a clear and strong timing guarantee to differentiate genuie and fake responses. 


\subsection{Face Verification}

We use \textit{Caffe}~\cite{jia2014caffe}, an open source deep learning framework, to train our neural network model used for face verification. The preliminary parameters are listed below: learning policy is set to ``multistep'', base learning rate is 0.1, gamma is 0.1, momentum is 0.9, weight decay is 0.0001 and the max iteration is 64000.

We first build a set of adversarial videos in order to train the model. These videos are made by recording the screen that is replaying the raw video. There are 4 different screens are recorded (Table~\ref{tb:screens}).

\begin{table}[tbh]
  \centering
    \caption{Four different screens.}
	\begin{tabular}{|c|c|c|c|}
	\hline
	& Screen & Resolution & Pixel Density\\
	\hline
	1 & HUAWEI P10 & 1920*1080 & 432(ppi) \\
	\hline
	2 & iPhone SE & 1136*640 & 326(ppi) \\
	\hline
	3 & AOC Monitor (e2450Swh) & 1920*1080 & 93(ppi) \\
	\hline
	4 & EIZO Monitor (ev2455) & 1920*1200 & 95(ppi) \\
	\hline
	\end{tabular}
	
	\label{tb:screens}
\vspace{-0.1in}
\end{table}

We take those frames in malicious videos as our negative samples, and take those raw videos' frames as positive samples. Besides, we bypass our timing verification to eliminate the mutual effect between these two verification algorithms. The experimental results are listed in the Table~\ref{tb:numbers}, which shows a zero false positive error with $99.2\%$ of accuracy rate.

\begin{table}
\vspace{-0.2in}
  \centering
  \caption{Experimental results of face verification.}
	\begin{tabular}{|c|c|c|c|c|}
	\hline
	& Training Ps & Training Ns & Testing Ps & Testing Ns\\
	\hline
	Total & 20931 & 20931 & 3000& 3000\\
	\hline
	Incorrect & 329 & 0 & 75 & 0\\
	\hline
	\end{tabular}
	
	\label{tb:numbers}
\vspace{-0.1in}
\end{table}

When applied with the testing data set, the accuracy is $98.8\%$. There are only 75 frames are incorrectly labeled, with all the negative samples labeled correctly. After analyzing these 75 frames, we found it may result from three reasons:
\begin{itemize}
\item Illumination. When the distance between face and screen is far and the environmental illumination is high, the captured response will be too obscure to be labeled correctly.
\item Saturation. Due to device limitations, video frames taken in dark scenarios, will have many saturated pixels, even having adjusted the sensitivity of optical sensors. As described in Section~\ref{sec:reflectionModel}, it is necessary to remove these saturated pixels to satisfy the formulas.
\item Vibration. Drastic head shaking and intensive vibration also fades our performance. Especially, we will not do so well on frames at the beginning and end of the captured video.
\end{itemize}

The above results showed that we can detect all the attacks with a small false negative error, which provides another security guarantee besides the response timing mentioned above.

\subsection{Evaluation on Robustness}
\label{sec:val:robust}

There are mainly two elements that could affect the performance of our proposed method: illumination and vibration. We have carefully designed six scenarios to further investigate their impacts. 
\begin{itemize}
	\item scenario 1: We instruct participants to stand in a continuous lighting room as motionless as possible. And the button was hidden during the first 15 seconds to let participants produce a long video clip.
	\item scenario 2: We instruct participants to take a subway train. The vibration is intermittent and lighting condition is changing all the time.
	\item scenario 3: We instruct participants to walk on our campus as they usually do during a sunny day.
	\item scenario 4: We instruct participants to hover under penthouses during a cloudy day.
	\item scenario 5: We instruct participants to walk downstairs at their usual speed in rooms.
	\item scenario 6: We instruct participants to walk down a slope outside during nights. 
\end{itemize}
We summarize the features of these scenarios in Table~\ref{tb:scene}.

\begin{table*}
  \centering
  \caption{Features of scenes.}
	\begin{tabular}{|c|c|c|c|c|c|c|}
	\hline
	& Scenario 1 & Scenario 2 & Scenario 3 & Scenario 4 & Scenario 5 & Scenario 6 \\
	\hline
	Illumination & good & varying & intense & normal & normal & dark\\
	\hline
	Vibration & no & intermittent & normal & normal & intense & intense\\
	\hline
	\end{tabular}
	
	\label{tb:scene}
\vspace{-0.1in}
\end{table*}

\begin{figure}[ht]
	\centering
  \begin{subfigure}{0.445\textwidth}
		\includegraphics[width=\textwidth]{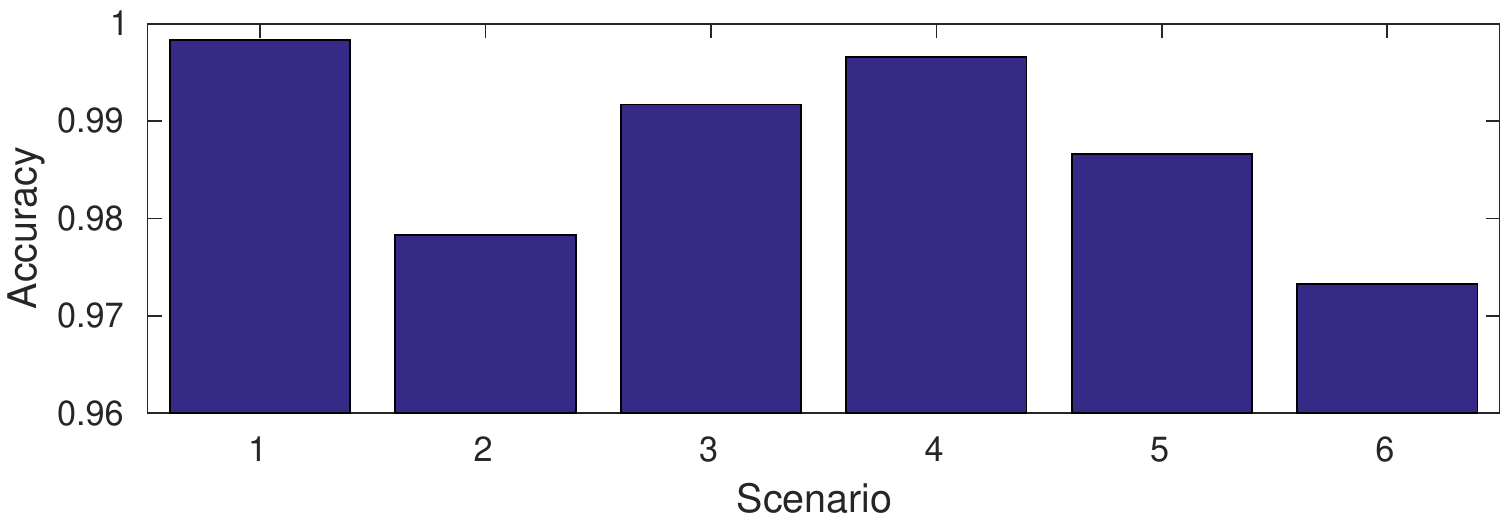}
  \end{subfigure}
  
  \caption{Performance on different scenarios.}
  \label{fig:scene}
\vspace{-0.2in}
\end{figure}

The results are shown on Fig~\ref{fig:scene}. In ideal environments (scenario 1), our method is perfect and the accuracy is high as $99.83\%$. In normal cases (scenario 4), our method is also excellent with the $99.17\%$ accuracy. And the sunlight (scenario 3) causes ignorable effects on the result, as long as the frontal camera does not face the sun directly. Comparing scenario 5 with 3, we infer the vibration causes more effect than the sunlight. Besides, dark is a devil (scenario 6) which reduces the accuracy to $97.33\%$, the lowest one. In our method, we cannot use the function of auto white balance (AWB) embedded in our devices, due to the fundamental requirement of our method. Adjusting the sensitivity of sensors, we just can limitedly reduce the effect of saturation, while keeping enough effectiveness. Limited by this constraint, the result is acceptable. For the complex case (scenario 2), the accuracy, $97.83\%$, is not bad. In this scenario, our device is being tested by many factors including unpredictable impacts, glare lamps and quickly changed shadows.

To further explore the impacts caused by vibrations, we built another experiment where we leveraged the six parameters generated by face tracking algorithm, and assembled them as a single value, $\nu$, to measure the intensity of vibration. The details are illustrated in Algorithm 2, where $\{T_j\}$ is the sequence of the transformation matrix (Section~\ref{sec:faceExtraction}) and $N$ is the number of frames.

\begin{figure}[ht]
	\centering
  \begin{subfigure}{0.23\textwidth}
		\includegraphics[width=\textwidth]{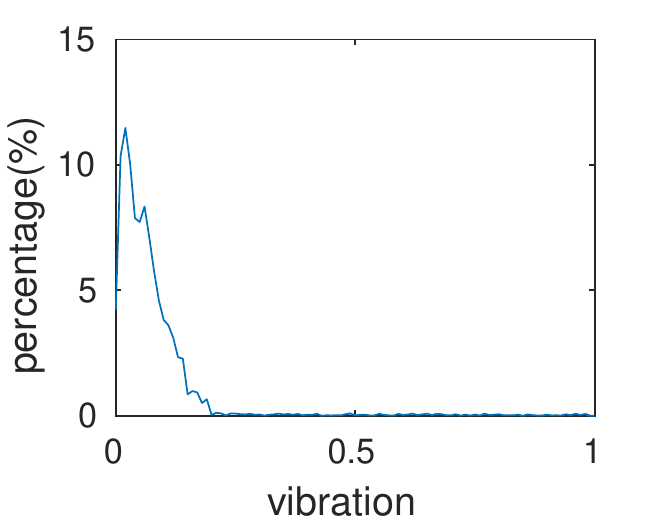}
    \caption{distribution}
    \label{fig:vib:ent}
  \end{subfigure}
	\begin{subfigure}{0.23\textwidth}
		\includegraphics[width=\textwidth]{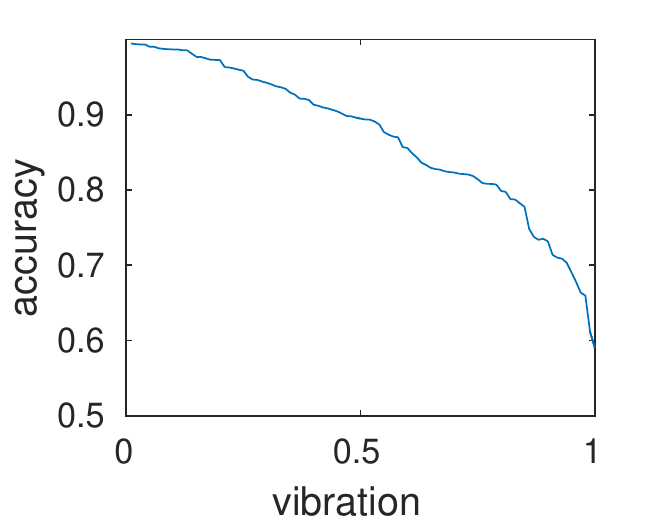}
    \caption{relation}
    \label{fig:vib:tot}
  \end{subfigure}
	\caption{Vibration effect.}
	\label{fig:vib}
\vspace{-0.25in}
\end{figure}

\begin{algorithm}[htb]
\label{alg:nu}
\caption{Algorithm to measure intensity of vibration.}
\begin{algorithmic}[1]

\INPUT{$\{T_{j}\}, N$}
\OUTPUT{$\nu$}
\For{$j=1 \ \ to \ \ N $}
\State  Extract face shifting, $(\alpha_{j},\beta_{j},\gamma_{j})$\;
\State  Extract face rotation, $(\iota_{j},\zeta_{j},\eta_{j})$\;
\EndFor
\State  Calculate mean values: $\bar{\alpha},\bar{\beta},\bar{\gamma},\bar{\iota},\bar{\zeta}$ and $\bar{\eta}$\;
\For{$i=j \ \ to \ \ N $}
\State  $\mu_{j}=\frac{\alpha_{j}}{\bar{\alpha}}+\frac{\beta_{j}}{\bar{\beta}}+\frac{\gamma_{j}}{\bar{\gamma}}+\frac{\iota_{j}}{\bar{\iota}}+\frac{\zeta_{j}}{\bar{\zeta}}+\frac{\eta_{j}}{\bar{\eta}}$\;
\EndFor
\State  $\nu=std(\{\mu_{j}\})$\;
\end{algorithmic}
\end{algorithm}
Fig~\ref{fig:vib:ent} shows the distribution of intensity. And Fig~\ref{fig:vib:tot} shows the relation between vibration intensity and accuracy. We divided all the intensity by the maximum value. From both figures, we can infer that vibration will produce side effects to our method and the most drastic vibration will reduce the accuracy to $60\%$. But, in general cases where the vibration is not that big, our method can perform very well. This means our method indeed is robust under normal vibration conditions. Particularly, when the intensity reaches $0.5$, we still hold $89\%$ accuracy.

In conclusion, our good robustness to vibration and illumination provides a good reliability and user experience. Besides, it excludes a potential attack scenario where adversary naively increases the vibration density.

\subsection{Computational and Storage Cost}
\label{cost}

The time costs of our method depend on concrete devices. If we run our method in the back-end server (say a laptop), the time needed to deal with 300 frames is less than 1 seconds, and the difference among time costs of our 3 steps is subtle. Here, we amplify this difference by running our method on a smartphone (Nexus 6) with a single core, and the resolution of all the frames are kept on 1920*1080. The time costs are shown in Table~\ref{tb:time}.

\vspace{-0.1in}
\begin{table} [ht]
  \centering
  \caption{Time cost of implementation on smartphone.}
  \resizebox{0.48\textwidth}{!} {
	\begin{tabular}{|c|c|c|c|c|}
	\hline
	Number of frames &50&100&200&300\\
	\hline
	Face extraction &6.11&11.70&20.12&28.63\\
	\hline
	Timing verification &0.03&0.05&0.11&0.22\\
	\hline
	Face verification &0.08&0.12&0.20&0.27\\
	\hline
	Total&6.22 (s) &11.87 (s) &20.43 (s)&29.12 (s)\\
	\hline
	\end{tabular}
  }

	\label{tb:time}
\end{table}
We discover that the most time-consuming step is the face extraction, which depends on algorithms we choose and the precision of face detection we want to achieve. The lower precision, the lower resolution of the input frame is needed and thus less time is needed. Particularly, if we shrink the input frame to half its size, the time will be reduced to about 1 second to extract the faces on 50 frames. The other way to reduce the time cost is leveraging the back-end server (the Cloud) in parallel, as we mentioned above. In practice, we keep the camera continuously recording the user's video and, parallely, sent ``.mp4'' files with each containing 30 frames recorded in one second, to our server through a 4G network (with about 12 Mbps of bandwidth in our experiment) for every second. Transferring one that file will consume 1.1 MB bandwidth. Once receiving a video, the server will perform our verifications on it and judge whether the user is benign. If the result of any second is negative, we regard this whole authentication session as a malicious attempt. Table~\ref{tb:time_cloud} demonstrates the time cost of this process. Compared with the implementation only on the smartphone, using cloud can significantly reduces the waiting and thus greatly improved user experiences.

\vspace{-0.1in}
\begin{table} [tbh]
  \centering
  \caption{Time cost of implementation using cloud.}
  \resizebox{0.48\textwidth}{!} {
	\begin{tabular}{|c|c|c|c|c|}
	\hline
	Number of frames &50&100&200&300\\
	\hline
	Recording in Front & 1.67&3.34&6.67&10\\
	\hline
	Verifying in Cloud &2.22&3.62&7.21&10.82\\
	\hline
	Time to Wait & 0.55 (s) & 0.28 (s) & 0.54 (s)& 0.82 (s)\\
	\hline
	\end{tabular}
  }
	\label{tb:time_cloud}
\vspace{-0.1in}
\end{table}

The storage space we need is the same as the size of captured videos, and the storage complexity is $O(NM)$. In real tests, 8.3Mb memory space is enough to store a video consisting 100 frames in JPG format.

\section{Related Works}
\label{sec:related}

Various liveness detection techniques have been proposed in the past decades. In this
section, we discuss differences between our method and those most relevant
previous studies. 

Our method could be categorized as a texture extraction method, according to the
classification in Chakraborty's survey~\cite{chakraborty2014overview}. The
traditional methods in this category mainly use various \textit{descriptors} to
extract features of images and pass features through a classifier to obtain the
final result. For instance, Arashloo et al.~\cite{benlamoudi2015face} used
multi-scale dynamic binarized statical features; Benlamoudi et
al.~\cite{benlamoudi2015face} used active shape models with steam and LBP; Wen
et al.~\cite{wen2015face} analyzed distortion using 4 different features, etc.
These methods work well under experimental conditions. But in our adversary model, the attacker can forge a perfect face that would defeat their approaches. 
In contrast, our method checks the geometric shape of the subject under authentication, and detect whether there are abnormal delays between responses and challenges. Even the adversary is technically capable of creating a perfect forged response, the time required in doing so will fail them.
Besides, previous works may fail due to the sub-optimal environmental conditions. However, our method is robust to that, as demonstrated in the evaluation part.       

Our method is also a challenge-response protocol. The traditional protocols are
based on human reactions. Comparing to them, our responses can be generated at the speed
of light. Li et al.~\cite{li2015seeing} proposed a new protocol that records the
inertial sensors' data while the user is moving around the mobile phone. If the data is consistent with the video captured by the mobile phone, the user is judged as a legitimate one. This method's challenge is the movement of mobile phone which is controlled by the user and measured by sensors. And the response is user's facial video which is also produced by the user. This method's security guarantee is based on the precise estimation of head poses. But we argued that the accuracy cannot be high enough in wild environment for two reasons: first, as mentioned by the authors, the estimation algorithm has about 7 degrees deviation; second, hand trembling produces side effect to the precision of the mobile sensors. In contrast, our approach is more robust, because, firstly, the challenges are fully out of attackers' control, and, secondly, our security guarantees are based on detecting the indelible delay, rather than the accurate estimation of the unstable head position.


Besides above methods, there is a close work published by Kim et
al.~\cite{kim2015face} who found that the diffusion speed is the distinguishing characteristic between real faces and fake faces. The reason is that the real face has more stereo shape which makes the reflection random and irregular. But this passive method will not work, when the environmental light is inefficient. 
From the figures shown in their paper, we can hardly distinguish the so-called binarized reflectance maps of malicious responses from legitimate responses, and these "vague" maps are fed to SVM for the final decision. So we argue that
this approach cannot defeat such attackers who have the ability to forge a perfect fake face. In contrast, our security guarantee is not only based on the stereo shape, but also the delay between responses and challenges. It's a very high bar for adversaries to forge a perfect response in such critical time. 
Another method leveraging reflection is proposed by Rudd et al.~\cite{rudd2016paraph}. The authors added two different polarization devices on the camera. And these devices impede the most of incoming light except the light in the particular direction. 
Comparing to this approach, our method does not require special devices and is more practical to use.

Specially, our work has overlaps with Andrew Bud's patent~\cite{iproovpatents} using also the light reflection to do the authentication. However, we focus on the security instead of functionality as they have done. Without our timing verification, as we demonstrated, there is no security guarantee and it is weak to defend against MFF attacks. And our work is independent from theirs.    

In general, compared with above relative works, Face Flashing is an active and effective approach with strong security guarantee on time. 


\section{Discussion}
\label{sec:discussion}

\bheading{Resilience to novel attacks.} An attack proposed by Mahmood et al.~\cite{mahmood2016accessorize} demonstrated that an attacker could 
impersonate the victim by placing a customized mask around his
eyes. 
Although such an attack can deceive
the state-of-the-art face recognition system, however, we believe it will be defeated by our method, as paper masks around the eyes can be easily detected by our neural network model in the verification of face (see Fig~\ref{fig:R_example:a} and~\ref{fig:R_example:c}).

\bheading{Challenge colors.} We used 8 different colors in our experiments. Considering the length of our challenge sequence, we believe these 8 colors are enough to provide a strong security guarantee. Because our security guarantee is achieved by detecting the \textit{delays}. If the adversary falsely infers one challenge, the \textit{delay} will be detected and her attempt will fail. Of course, we can easily increase the space of the challenge sequences by using the striped pictures with a more sophisticated algorithm.

\bheading{Authentication time.} Our method needs a few seconds to gather enough
responses for authentication. As we mentioned in data collection's part, 3 seconds is a reasonable default setting. In this period, we can choose sufficient responses with high quality, and the user can complete the instructed expression. Essentially, 1 second is enough for our method to finish the work, but the user will be in a hurry.

\bheading{Other applications of our techniques.} One interesting application of our method is to improve the accuracy of state-of-the-art face recognition algorithms by distilling the personal information contained in the geometric shape. We believe the shape is unique. The
combined method will have stronger ability to prevent advanced future
attacks.

\section{Limitations}
\label{sec:limitation}
The silicone mask may pass our system. But, this mask is hard to be fabricated (3D printed) due to the reasons mentioned in Section~\ref{sec:attackingTree}. And our system has the potential to defeat it completely, owing to our unique challenges: lights of different wavelength (colors). According to previous studies~\cite{weyrich2005measurement}, light reflected from human skin has an ``albedo curve'', the curve depicting reflectance of different wavelengths. Therefore, the reflections from different surfaces can be distinguished by discernible albedo curves, which enables Face Flashing to recognize attackers wearing such ``soft'' masks. However, this technique is sophisticated and deserves another paper.

Even though we raise the bar of the attacks, we cannot totally neutralize adversaries' advantages coming from super devices. They still have a chance to pass our system, if they somehow use an ultrahigh-speed camera (FASTCAM SA1.1 with 675000fps), an ultrahigh-speed screen in the similar level (says with 100000Hz), and the solution to reduce the transmission and buffering delays. In this situation, adversaries can instantly forge the response to every challenge with small delays and subtle variance, so our protocol will fail. However, this attack is expensive and sophisticated. On the other side, we can mitigate this threat, to some extent, by flashing more finely striped challenges (or chessboard-like patterns), but, with better screen and camera.

\section{Conclusion}
\label{sec:conclusion} 

In this paper, we proposed a novel challenge-response protocol, Face Flashing,
to defeat the main threats against face authentication system---the \textit{2D
dynamic} attacks. We have systematically analyzed our method and illustrated that our
method has strong security guarantees. We implemented a prototype that does verifications both on time and the face. We have demonstrated
that our method has high accuracy in various environments and is robust to
vibration and illumination. Experimental results prove that our protocol is
effective and efficient.


\section*{Acknowledgment}
We thank our shepherd Muhammad Naveed for his patient guidance on improving this paper, and anonymous reviewers for their insightful comments. We also want to thank Tao Mo and Shizhan Zhu for their supports on the face alignment and tracking algorithms. This work was partially supported by National Natural Science Foundation of China (NSFC) under Grant No. 61572415, Hong Kong S.A.R. Research Grants Council (RGC) Early Career Scheme/General Research Fund No. 24207815 and 14217816. 




%

\bibliographystyle{IEEEtranS}
\bibliography{main}

\end{document}